\documentclass[conference,compsoc]{IEEEtran}
\IEEEoverridecommandlockouts

\usepackage{cite}
\usepackage{amsmath,amssymb,amsfonts}
\usepackage{algorithmic}
\usepackage{graphicx}
\usepackage{textcomp}
\usepackage[table]{xcolor}
\usepackage{xfp}
\usepackage{multirow}
\usepackage{array}
\usepackage{booktabs}
\usepackage{threeparttable}
\usepackage{dblfloatfix}
\usepackage{subcaption}
\usepackage{url}
\def\BibTeX{{\rm B\kern-.05em{\sc i\kern-.025em b}\kern-.08em
    T\kern-.1667em\lower.7ex\hbox{E}\kern-.125emX}}
\newif\ifinternalnotes
\internalnotesfalse
\begin{document}

\title{NeuroShield: A Device-Agnostic Foundation Model for EEG Authentication}
\author{
\IEEEauthorblockN{Matin Fallahi\IEEEauthorrefmark{1}, Patricia Arias-Cabarcos\IEEEauthorrefmark{2}, and Thorsten Strufe\IEEEauthorrefmark{1}}
\IEEEauthorblockA{\IEEEauthorrefmark{1}Karlsruhe Institute of Technology (KIT), Karlsruhe, Germany}
\IEEEauthorblockA{\IEEEauthorrefmark{2}Joint Research Centre, European Commission}
\thanks{Code and pretrained model: https://github.com/kit-ps/NeuroShield-FM}
}

\ifinternalnotes
\input{Sections/00_internal_reviewer_notes}
\clearpage
\fi

\maketitle

\begin{abstract}
A central challenge in EEG authentication is that models are typically tied to the acquisition settings in which they are trained. In particular, variations in headset hardware, channel layout, and signal duration create heterogeneous recordings that existing models are not designed to handle, causing each new headset or dataset to be treated as a separate model-development problem. This fragmentation limits multi-dataset learning, hinders knowledge transfer, and reduces model reusability.
To address this limitation, we present NeuroShield, a reusable foundation model for EEG authentication that learns identity-discriminative embeddings from variable-channel
and variable-length EEG recordings through a dual-stage
transformer architecture. We pretrain NeuroShield on three public EEG datasets
comprising 15{,}762 subjects and 28{,}116 sessions, and
evaluate transfer on two unseen downstream datasets.
Our evaluations show that, after fine-tuning, NeuroShield reduces equal error rate by 0.44--8.06 percentage points relative to the state of the art.
NeuroShield further generalizes to segments longer than those seen during training and operates across channel layouts not encountered during pretraining.
These results establish NeuroShield as a reusable and adaptable EEG identity encoder across heterogeneous recording settings. We release NeuroShield as open source to support reproducibility and community adoption.
\end{abstract}


\section{Introduction}

\begin{figure}
    \centering
    \includegraphics[width=0.90\linewidth]{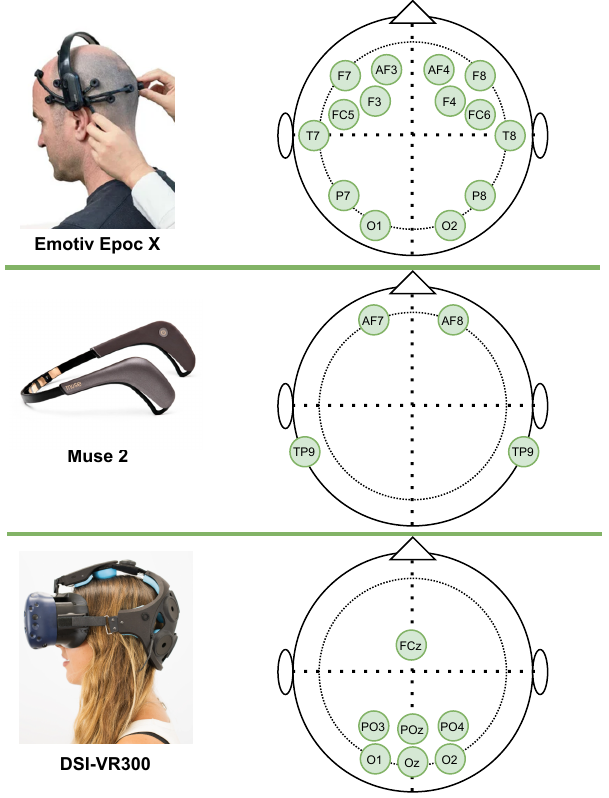}
    \caption{Examples of heterogeneous EEG headset layouts. Differences in channel number, scalp coverage, and electrode placement motivate flexible-input authentication models.}
    \label{fig:heterogeneous_headset}
\end{figure}

EEG (electroencephalogram) authentication offers a distinctive route to hands-free, continuous, and revocable verification as it uses brain activity recorded by wearable devices rather than static, externally visible traits~\cite{fidas2023review}.
As EEG acquisition
moves beyond controlled laboratory settings toward
wearable and consumer-grade devices, the central
challenge is no longer only whether EEG contains
subject-discriminative information, but whether
authentication models can preserve this information
across recording conditions~\cite{chan2018challenges,
fallahi2026advancing}. Current EEG authentication
research remains fragmented across datasets, headsets,
and channel layouts~\ref{fig:heterogeneous_headset}, so models are usually developed
for a single recording setup and transfer poorly
beyond it~\cite{chaurasia2024neuroidbench,
fallahi2026advancing}. Consequently, the field continues
to rely on dataset-specific training pipelines rather
than reusable models that can accumulate knowledge
across studies and support deployment in heterogeneous
settings.

The technical source of this limitation is that
most EEG authentication models are designed around
a fixed input
structure~\cite{hernandez2022eeg,al2025eeg,
fallahi2026advancing}. They typically assume a
predefined channel set, a stable channel order,
and a fixed temporal window. These assumptions are
difficult to satisfy across datasets because EEG
headsets differ in scalp coverage and channel
layout, while acquisition protocols and deployment
settings differ in how much usable signal is
available for verification. Existing studies usually
handle these differences by adapting the model
input separately for each dataset--and, ultimately, for each hardware setup or potential deployment setting--for example by changing the expected channel configuration,
restricting the input to selected channels, or
fixing signal durations to those found in the various datasets~\cite{chaurasia2024neuroidbench,
fallahi2023brainnet,debie2021session}. This can
produce strong results in tailored academic experiments,
but such models do not transfer reliably when electrode positions, channel layout, number of channels, or signal timing change.
As a result, direct inference on new datasets or deployment settings remains limited,
performance can degrade when the input structure changes,
and useful spatial or temporal information contained in EEG data may be discarded.
As a result, each new dataset or
headset remains a separate modeling problem rather
than contributing to a reusable authentication
model.

To address this limitation, we introduce NeuroShield, a foundation model for EEG authentication
that supports flexible inputs, including varying numbers of channels, diverse channel placements,
and variable signal durations, through a dual-stage transformer architecture with temporal and
spatial positional encoding mechanisms (Section~\ref{sec:design}).
This flexible input design allows NeuroShield to be pretrained on three
public EEG datasets comprising 15{,}762 subjects and 28{,}116 sessions
spanning multiple channel layouts and acquisition systems (Section~\ref{sec:methodology}).
We evaluate the advantage of the architecture
through cross-dataset
transfer, as well as temporal and spatial flexibility evaluations
(Section~\ref{sec:eval-main-claims}), then investigate
the effects of training-data composition and
deployment-oriented robustness properties
through further analyses
(Sections~\ref{sec:training-data} and \ref{sec:missing-channels}).
Finally, we compare NeuroShield with related studies and
biometric standards, then conclude
with limitations, availability, and future directions
(Sections~\ref{sec:comparison},
and \ref{sec:conclusion}).

Our main contributions are as follows:
\begin{itemize}
    \item We introduce NeuroShield as a reusable
    foundation-model for EEG authentication across heterogeneous
    channel layouts and signal durations, instantiated through
    a dual-stage transformer architecture
    (Section~\ref{sec:design}).

    \item We demonstrate that  NeuroShield supports
    cross-dataset zero-shot transfer and, after fine-tuning,
    reduces downstream authentication error by 0.44\%--8.06\%  points relative to the
    reproduced benchmark baseline on the main downstream
    evaluations
    (Section~\ref{sec:eval-main-claims}).

    \item We evaluate NeuroShield's temporal and spatial input flexibility, showing that it operates across variable temporal window lengths, including windows longer than those used during training, and generalizes to new channel layouts
    (Sections~\ref{sec:eval-temporal} and \ref{sec:eval-spatial}).

    \item We provide deployment-oriented analyses,
    showing that NeuroShield can tolerate about
    10\%--30\% verification-time channel loss, with lower degradation for headsets with a higher number of channels,
     and complementary training-data analyses
    showing that multi-session diversity is more valuable
    than simply adding more single-session samples
    (Sections~\ref{sec:training-data} and \ref{sec:missing-channels}).

    \item We make the NeuroShield foundation model and its
    evaluation pipeline open source on GitHub to support future
    EEG-authentication research and reproducible cross-dataset
    comparison.
\end{itemize}

\section{Preliminaries}

This section introduces the background most relevant
to NeuroShield, including biometric authentication,
EEG as an authentication signal, transformer-based
sequence modeling, and the threat model used in our
evaluation.

\subsection{EEG for Biometric Authentication}
Authentication is the process of verifying whether a user is the legitimate holder of a claimed identity. In biometric authentication, this decision is based on physiological or behavioral characteristics rather than knowledge-based credentials or physical tokens. A biometric authentication system typically involves two phases: enrollment and verification~\cite{jain2004introduction}. During enrollment, biometric samples are collected from a user and transformed into a stored template. During verification, a newly collected sample from the claimed user is verified against the enrolled template to decide whether the authentication attempt should be accepted or rejected. In real-world biometric use, enrollment is performed first, and verification follows after a time interval, such as days or weeks later. Many prior EEG authentication studies, however, evaluate enrollment and verification within a single recording session~\cite{cai2023aitst,arias2023consumer,fallahi2026beyond}.

Both enrollment and verification begin with collecting a biometric sample, followed by preprocessing to reduce noise and improve data quality~\cite{jain2004introduction}. Relevant features are then extracted and compared against the enrolled template to determine whether the sample belongs to the claimed user. Traditionally, biometric systems rely on handcrafted features designed using domain knowledge, which are subsequently used to train a classifier that distinguishes between genuine users and impostors~\cite{arias2023consumer}. More recently, deep learning approaches have become dominant. Instead of manually designing features, a neural network is trained to learn subject-discriminative feature representations directly from the data~\cite{fallahi2026advancing}. The resulting embedding vectors capture identity-related characteristics and can be efficiently compared using similarity or distance measures, such as cosine similarity, to perform authentication.

EEG is a non-invasive method for
recording brain activity through channels connected to electrodes on the scalp, termed scalp channels. Since
its first human recordings by Hans
Berger~\cite{lemke2024hans}, EEG has become a standard
tool in neuroscience and clinical neurophysiology. The
recent advances in hardware have
pushed EEG beyond specialized clinical settings toward
wireless, wearable, and consumer-grade
devices~\cite{sawangjai2020consumer}. This shift has
motivated interest in EEG as a biometric signal for
user authentication, since it can support hands-free,
implicit, and continuous authentication while relying on
a live physiological signal that is more difficult to
observe directly than external
traits~\cite{fallahi2026advancing,fidas2023review}.
At the same time, practical EEG authentication must cope
with variation in channel layout, recording hardware,
and usable signal duration across datasets and devices.
These properties motivate models that can learn reusable
subject representations across heterogeneous EEG settings rather
than being tied to one fixed physical setup and acquisition protocol.

EEG authentication studies acquire signals under different recording tasks\cite{fidas2023review,bidgoly2020survey}. These include resting-state recordings, in which users keep their eyes open or closed without responding to explicit stimuli; event-related paradigms, in which short EEG responses are recorded around external stimuli\cite{al2025eeg}; and cognitive tasks, such as memory or attention tasks, that elicit task-dependent brain activity over longer intervals\cite{maiorana2021taskindependent,fallahi2026advancing}. These paradigms differ in the amount and structure of usable EEG data. For example, event-related settings often provide short event-centered windows, whereas resting-state and cognitive protocols can provide longer continuous segments ranging from a few seconds to tens of seconds in prior EEG-authentication studies\cite{al2025eeg,maiorana2021taskindependent,fallahi2026advancing}.

\subsection{Transformer-based signal encoder}
Transformers were introduced by Vaswani et al.~\cite{vaswani2017attention}
for machine translation and other sequence modeling tasks, as an alternative
to recurrent and convolutional sequence models. Their central mechanism is
self-attention, which represents an input as a sequence of tokens and updates
each token representation by aggregating information from other tokens in the
sequence. This allows the model to capture long-range dependencies without
processing the sequence strictly step by step. However, self-attention alone
does not encode token order. Transformers therefore require positional information
so that the model can distinguish tokens according to their position in the
sequence. This positional information can be provided through absolute
positional encodings, learned positional embeddings, or relative-position
mechanisms. Among these, relative-position mechanisms encode position through
distances between tokens, making them better suited when a model needs to
extrapolate to longer sequences. Attention with Linear Biases
(ALiBi)~\cite{press2022train} implements this idea by adding fixed
distance-dependent linear biases to the attention scores, encouraging
the model to account for relative token distance while having been shown
to support extrapolation to longer sequence lengths.

Although transformers were originally introduced for
language, the same principle is also effective for time-series
signals: a signal can be divided into ordered tokens, and
self-attention can model dependencies between these tokens
over time~\cite{zerveas2021transformer,zhou2021informer}.
EEG signals for authentication can therefore be segmented and
represented as tokens, enabling transformer models to exploit
information flow across tokens and capture temporal
dependencies. Once temporal modeling is achieved with
transformers, their ability to process variable-length inputs
can be leveraged to build flexible-input models for EEG
authentication.

\subsection{Threat Model}

This work introduces NeuroShield for the biometric
matching layer. The objective is to investigate whether
its learned EEG embedding space supports reliable
subject verification across devices, temporal configurations, and
channel layouts, rather than to evaluate a complete
deployed authentication system. Consistent with
standards-oriented biometric evaluation, including
NIST SP 800-63B and ISO/IEC
30107-1~\cite{nist80063b2025,iso30107_1_2023}, we
focus on zero-effort impostor attempts as the
baseline setting for characterizing biometric
matching errors. In this setting, an impostor
submits their own EEG signal for comparison against
another subject's enrollment template.

We assume that the adversary knows the authentication
procedure and may have access to comparable EEG
acquisition hardware, but can only interact with the
system through the normal verification interface.
The adversary cannot modify the model parameters,
preprocessing pipeline, enrollment procedure, stored
templates, or biometric database. Under this threat
model,
the reported verification results quantify NeuroShield's matching errors under standard zero-effort impostor conditions.

Accordingly, the evaluation isolates the core
biometric matching challenge in EEG authentication:
whether another user's own EEG embedding is similar
enough to the target user's enrollment template to
be falsely accepted. In addition, all subjects in
the evaluated datasets perform the same acquisition
tasks, so observing the task protocol does not
provide target-specific information beyond what is
already present in the evaluation setup. EEG signals are also not
externally visible traits; to the best
of our knowledge, no prior work shows that a human
can intentionally imitate a target user's brain activity
in a way that provides an advantage in matching the target's
EEG template, unlike observable behaviors such as gestures.

\section{Related Work}

EEG has been studied as a biometric signal for user
authentication for more than a decade and has shown that
brain activity can contain subject-discriminative
information~\cite{marcel2007brainwaves,fidas2023review,gui2019survey}.
Early approaches mainly relied on handcrafted EEG
features combined with subject-specific binary
classifiers~\cite{marcel2007brainwaves,bidgoly2020survey}.
More recent work has shifted toward deep models
that learn discriminative representations directly
from the signal~\cite{maiorana2021taskindependent,
fallahi2023brainnet}. In particular, embedding-based verification
pipelines map enrollment and verification samples into a shared
representation space and compare their embeddings using a similarity
metric. This is attractive because the model learns a general embedding
function rather than a closed-set classifier over the training subjects.
As a result, a new user can be enrolled by storing their embedding template,
and later verification samples from that user can be compared against this
template without retraining the full
model~\cite{maiorana2021taskindependent,fallahi2023brainnet}. Together,
these studies show that EEG contains subject-discriminative information
suitable for user authentication.

However, most prior EEG authentication studies are
still developed and evaluated in isolated
dataset-specific settings~\cite{maiorana2021taskindependent,
bidgoly2022universal,zhao2026brainwave,fallahi2026advancing}.
Even when multiple datasets are considered, models
are typically trained and evaluated separately from
scratch rather than transferring knowledge across
datasets in a reusable way for subsequent
work~\cite{debie2021session,fallahi2023brainnet,
chaurasia2024neuroidbench,arias2023consumer}. As a
result, the literature provides many strong
dataset-level systems, but little support for
cross-dataset reuse.

A key reason is the strong heterogeneity across EEG collections. Datasets differ substantially in channel layout, including both the number of available channels and their scalp coverage across devices and vendors. Even when headsets are guided by common standards such as the 10--20 system~\cite{jurcak2007ten}, they may follow different variants or use different channel subsets, which still introduces substantial spatial variation across datasets. Datasets also differ in the amount of EEG available at verification time. For example, prior work has used different task protocols that produce EEG segments of different lengths: 5~s segments~\cite{maiorana2021taskindependent}, 0.8--4~s trial segments~\cite{debie2021session}, and 1~s samples aggregated into longer verification attempts~\cite{fallahi2026advancing}. In practice, these differences are often handled by adapting each dataset to a fixed model input, which helps explain the continued reliance on architectures such as CNN-based pipelines~\cite{maiorana2021taskindependent,fallahi2023brainnet,fallahi2026advancing}. RNN-style alternatives have shown some promise toward temporal flexibility, but still fall short in performance~\cite{fallahi2026advancing}. Such input constraints make knowledge transfer difficult, since it cannot be assumed that the results on a model learned under one recording setting carries over to another with different channel layouts or authentication durations.

Transformers offer a promising route toward flexible EEG modeling
because they can process input sequences with varying numbers of
tokens while using self-attention to model relationships among those
tokens. This property is important in our setting, where an EEG authentication
model must learn from heterogeneous datasets with varying channel layouts and
input durations and generalize to unseen headset configurations.
While prior EEG user-recognition studies have demonstrated
promising results when using transformer architectures as core model building
blocks~\cite{du2022etst,cai2023aitst,shao2026clt,
li2025enhancing}. However, these works mainly focus
on improving authentication or recognition performance
within a given isolated dataset setting,
rather than on
supporting input flexibility across heterogeneous EEG
collections. As a result, they are still
designed and evaluated under fixed windowing and
channel configurations. Even transformer studies that consider
multiple datasets ultimately train and evaluate
separate dataset-specific models rather than reusing
one model across heterogeneous inputs~\cite{shao2026clt,li2025enhancing}.

Taken together, prior EEG authentication studies show that EEG contains
subject-discriminative information and that existing methods can perform
well when training and evaluation use the same acquisition setting. However,
these approaches are trained from scratch for each headset, dataset, or task
because they assume a fixed channel layout and signal duration. This design
limits reuse when channel availability, electrode positions, or verification-window
length change, and also prevents them from benefiting from other datasets to improve
performance. Addressing this limitation requires a flexible-input authentication
architecture that can learn across heterogeneous EEG collections and transfer to
unseen acquisition settings. Although transformers can handle variable-length token
sequences, prior EEG authentication and user-recognition studies have not used this
property to build a reusable verification encoder for such input variation.

\begin{figure*}
    \centering
    \includegraphics[width=1\linewidth]{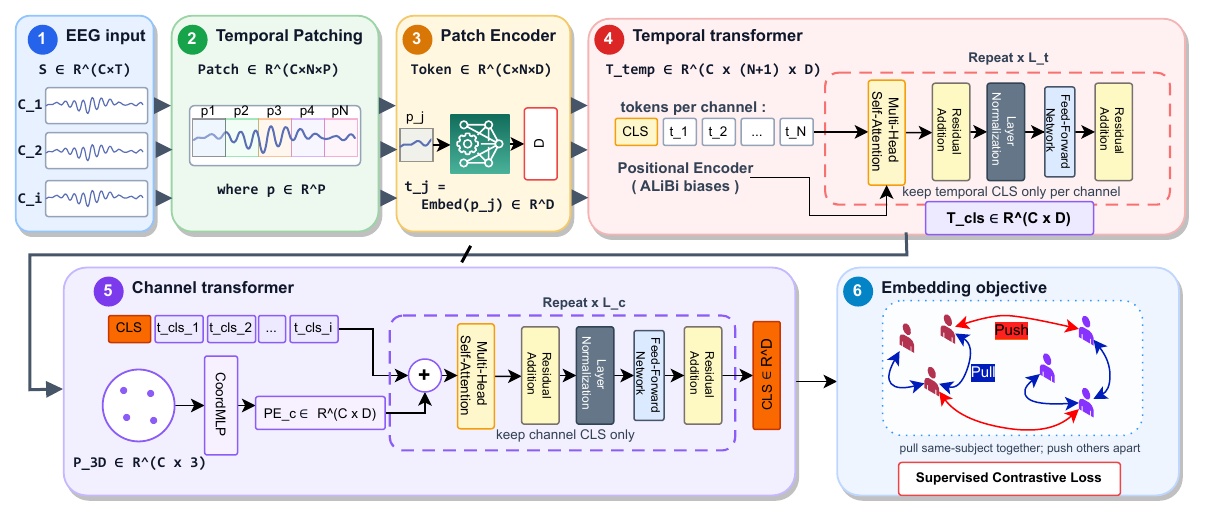}
    \caption{\small \textbf{Overview of the NeuroShield architecture.} The input EEG segment is patchified along time and encoded into patch tokens. A temporal transformer with fixed ALiBi biases summarizes each channel into a temporal CLS token, and these channel summaries are fused with geometry-aware embeddings derived from 3D channel positions in a channel transformer. The resulting segment embedding is trained with supervised contrastive loss to increase subject separability in the learned metric space. To simplify the illustration, steps~2--4 are shown for a representative channel pathway, but in the actual model these operations are applied in parallel to all channels.}
    \label{fig:Overview}
\end{figure*}

\section{NeuroShield Design}
\label{sec:design}

NeuroShield is designed as a foundation model for
EEG authentication: a reusable encoder that maps
heterogeneous EEG recordings into a shared
subject-discriminative feature representation.

The central design requirement is to produce a stable
authentication representation even when EEG samples
differ in available channel set, channel layout,
and temporal length.

To facilitate robust performance for variations in both signal duration and channel layout, we separate temporal modeling and spatial aggregation. Rather than using a fixed-input architecture that jointly models all channels and temporal samples as a single input, which limits knowledge transfer to headsets with different layouts or authentication settings with different signal durations, NeuroShield uses a dual-stage transformer architecture. \textbf{In the first stage}, the model learns a location-independent temporal encoding for each available EEG channel. Concretely, each single-channel waveform is divided into temporal patches, encoded as a sequence of patch tokens, and passed through a temporal transformer that extracts within-channel temporal features. The resulting temporal classification token (CLS) token represents the channel-level embedding. Because the temporal transformer operates on a sequence of patch tokens, it can process different numbers of tokens and therefore handle EEG segments of different lengths. \textbf{In the second stage}, the channel-level embeddings are combined across channels using geometry-aware channel embeddings derived from electrode positions. This allows NeuroShield to capture spatial relationships between electrodes without assuming a fixed channel order or a fixed channel layout.

Following the workflow shown in Fig.~\ref{fig:Overview}, NeuroShield processes an EEG segment through the following stages:

\textbf{EEG input:} Downstream EEG settings may differ in both channel availability and recording duration. NeuroShield therefore represents each EEG segment as \(S \in \mathbb{R}^{C \times T}\), where (C) denotes the number of available channels and (T) denotes the number of temporal samples, and is designed to support variation in both (C) and (T). Here, temporal samples refer to the data points recorded for each channel, whose number depends on the segment duration and the sampling rate. To keep the sampling rate fixed across datasets, we resample all segments to 500 samples per second. Therefore, (T) varies with the segment duration; for example, a 3-second EEG segment contains 1500 temporal samples per channel.

\textbf{Temporal Patching:} The temporal transformer requires a sequence of temporal units as input, so that it can first encode local temporal patterns and then model longer-range temporal dependencies through self-attention. At this stage, the data from each channel is represented as a time-continuous waveform. NeuroShield therefore divides each channel waveform into smaller fixed-length temporal patches of length (P), following time-series transformer designs that use multi-timestamp inputs to preserve local temporal patterns~\cite{nie2023patchtst,wang2024medformer}. This produces \(S_{\text{patch}} \in \mathbb{R}^{C \times N \times P}\), where \(N\) is the number of temporal patches.

\textbf{Patch encoder:} After patchification, each temporal patch is still a raw signal fragment. Although patchification defines the temporal unit that each token should cover, the transformer cannot operate directly on raw sample values; it needs learned token representations in a shared latent space so that relationships between patches can be modeled through attention. NeuroShield therefore uses a patch encoder to map each patch \(p_j \in \mathbb{R}^{P}\) into a \(D\)-dimensional embedding, yielding \(T_{\text{patch}} \in \mathbb{R}^{C \times N \times D}\). These learned patch tokens are then passed to the temporal transformer.

\textbf{Temporal transformer:} At this point, each channel is represented as a sequence of patch tokens. Authentication-relevant temporal patterns can be distributed across the segment and can also depend on their temporal order. Meanwhile, the number of patches may vary because the model is designed to support temporal input flexibility. To handle this, NeuroShield needs a compact temporal summary. Therefore, for each channel, a temporal classification token (CLS) is prepended to the patch-token sequence, forming \(T_{\text{temp}} \in \mathbb{R}^{C \times (N+1) \times D}\), and self-attention is applied independently within each channel to model long-range temporal dependencies.
To preserve temporal order while remaining compatible with variable sequence lengths, the temporal transformer uses fixed ALiBi biases, which provide a relative position signal and support input-length extrapolation~\cite{press2022train}. After $L_t$ temporal transformer layers, only the temporal CLS token is retained from each channel, resulting in \(T_{\text{cls}} \in \mathbb{R}^{C \times D}\). Here, $L_t$ denotes the temporal depth and is selected during hyperparameter tuning.

\textbf{Channel transformer:} After the temporal transformer, each channel is summarized into one temporal token. The next step is to aggregate these channel-level tokens into a final representation of the input. The challenge is that, to preserve channel flexibility, the model should not assume a fixed input channel set or a fixed channel order. To handle this, NeuroShield uses the 3D channel coordinates as input to a learned coordinate encoder, which maps each channel position to a geometry-aware positional embedding \(PE_c \in \mathbb{R}^{C \times D}\). These embeddings are added to the channel tokens, and a channel CLS token is placed at the beginning of the sequence, giving \(T_{\text{chan}} = [\mathrm{CLS}_c,\; T_{\text{cls}} + PE_c] \in \mathbb{R}^{(C+1) \times D}\). Self-attention is then applied across channels to aggregate spatially distributed subject information. The final channel CLS token is used as the summary of the full EEG input and the representation of the user.

\textbf{Embedding objective:} After the channel transformer, the full EEG segment is summarized into one final channel CLS token. This token is an architectural summary of the input, but it is not yet a meaningful authentication representation by itself. NeuroShield therefore trains this output as a subject-discriminative embedding: samples from the same user are encouraged to be close in the embedding space, while samples from different users are encouraged to be separated. The resulting vector is used as the segment-level representation ($e \in \mathbb{R}^{D}$).

We optimize this representation using supervised contrastive loss (SupConLoss)~\cite{khosla2020supervised}. SupConLoss uses subject labels to define positive and negative relations within each batch: for each anchor embedding, embeddings from the same subject act as positives, while embeddings from other subjects act as negatives. Compared with pairwise or triplet-based objectives, this batch-wise supervision uses multiple same-subject and different-subject relations simultaneously, encouraging compact same-subject clusters and clearer separation between subjects. In our implementation, SupConLoss is combined with a Multi-Similarity Miner~\cite{wang2019multisimilarity} to emphasize informative relations within each batch. This design choice is further supported by prior work~\cite{fallahi2026advancing}, which benchmark metric-learning objectives for large-scale EEG biometrics and identify SupConLoss as the best-performing feature-extraction objective.

\begin{table*}[t]
\centering
\small
\setlength{\tabcolsep}{5pt}
\renewcommand{\arraystretch}{1.15}
\begin{threeparttable}
\caption{Summary of the EEG datasets used in this work.}
\label{tab:datasets}
\begin{tabular}{l l r c r p{4.2cm}}
\toprule
Split & Dataset & \#Subjects & \#Sessions & \#Channels & Hardware \\
\midrule

\multirow{3}{*}{Train}
& TUH-EEG~\cite{harati2014tuh} & 14987 & 26836 & 17--22 & Nicolet systems \\
& DVS-rsEEG~\cite{ds005385:1.0.3,getzmann2024resting} & 608   & 816     & 64      & BrainVision BrainAmp DC \\
& PREDICT~\cite{ds005486:1.0.1,seminowicz2020novel,chowdhury2025predicting} & 167   & 464     & 63      & Brain Products actiCHamp Plus \\
\midrule

\multirow{2}{*}{Evaluation}
& PEERS~\cite{ds004395:2.0.0,kahana2024penn} & 345   & 6007  & 89     & EGI GSN / BioSemi ActiveTwo \\
& FRC-EEG~\cite{albasri2019eeg} & 30    & 120    & 14     & Emotiv Epoc X \\
\bottomrule
\end{tabular}
\end{threeparttable}
\end{table*}

\section{Experimental Protocol}
\label{sec:methodology}

To evaluate NeuroShield as a reusable foundation model for EEG authentication across different headsets, channel layouts, and segment durations, we implement an EEG biometric evaluation pipeline. This pipeline first applies the preprocessing procedure used in prior state-of-the-art EEG authentication work, described in Appendix~\ref{app:preprocessing}. It then uses datasets that cover heterogeneous recording conditions, including different acquisition hardware, channel configurations, and session structures. Within this pipeline, NeuroShield serves as the feature extractor that maps each EEG segment into an embedding representation. Authentication scores are computed using cosine similarity between enrollment and verification embeddings, following prior EEG authentication work~\cite{fallahi2026advancing}, and performance is measured using standard biometric metrics. Since reusable models may still require different degrees of target-dataset adaptation, we also specify the fine-tuning settings investigated in this work. Finally, we reproduce a state-of-the-art fixed-input baseline to provide a reference point for evaluating whether NeuroShield improves over conventional dataset-specific EEG authentication pipelines.

\subsection{Datasets}
To develop and evaluate NeuroShield as a reusable EEG
authentication encoder, we require two groups of datasets:
source datasets for foundation-model training and target datasets
for downstream transfer evaluation. This separation allows us
to test whether knowledge learned from one set
of EEG headsets and channel layouts transfers to
unseen acquisition settings. Therefore, the target datasets are
collected with hardware that is not used during
foundation-model training. Since EEG authentication depends on subject
information that remains stable from enrollment to later
verification, both groups should include multi-session recordings. At
the same time, because large multi-session EEG datasets
are predominantly collected with medical- or research-grade systems,
we include a consumer-grade target dataset to assess
the extent to which such transfer can benefit
smaller wearable EEG headsets. Table~\ref{tab:datasets} summarizes
the datasets used in this work.

For foundation-model training, we select large public EEG datasets
with many subjects who have repeated multi-session recordings, since
this is important for learning subject representations that remain
stable across sessions. The training split includes
TUH-EEG~\cite{harati2014tuh},
DVS-rsEEG~\cite{ds005385:1.0.3,getzmann2024resting}, and
PREDICT~\cite{ds005486:1.0.1,seminowicz2020novel,chowdhury2025predicting}.
Among these, TUH-EEG provides 14{,}987 subjects and 26{,}836
recordings, making it the largest public multi-session EEG
corpus used in this study. DVS-rsEEG and PREDICT contribute additional multi-session data collected with different research-grade hardware. Across the three training datasets, NeuroShield is exposed to heterogeneous channel
configurations, ranging from 17 to 64 channels per recording
and covering 69 unique channels across seven distinct channel
layouts. These datasets also span multiple acquisition systems,
including Natus Medical Nicolet EEG systems~\footnote{Recordings
were acquired using several generations of Natus Medical Incorporated's
Nicolet EEG recording technology.}, BrainVision BrainAmp DC, and
Brain Products actiCHamp Plus.

For downstream evaluation, we use PEERS~\cite{ds004395:2.0.0,kahana2024penn}
and FRC-EEG~\cite{albasri2019eeg}. PEERS provides 345 subjects,
6{,}007 sessions, and 89-channel recordings, and serves as
the main multi-session evaluation target for comparison against
recent EEG-authentication results~\cite{fallahi2026advancing}. FRC-EEG
contains 30 subjects, 120 sessions, and 14 channels recorded
with an Emotiv Epoc X headset, and is used to evaluate
transfer to a smaller consumer-grade target setting. Since both
evaluation datasets are collected with hardware that differs from
the training datasets, they allow us to examine cross-device
transferability.

All datasets are partitioned by subject identity into train,
validation, and held-out test splits to prevent samples from
the same subject appearing in multiple splits. For foundation-model
training, the source datasets are used only through their
training and validation subjects for NeuroShield hyperparameter tuning
and model training. Their held-out test subjects are reserved
for final evaluation on the source-data distribution, which we
refer to as FM Train-Test. The downstream target datasets,
PEERS and FRC-EEG, are split independently from the source
datasets and from each other. Their training and validation
splits are used only when a downstream setting requires
target-dataset adaptation or fine-tuning, whereas their held-out test
splits are reserved for final transfer evaluation. This split
design ensures that target-dataset test subjects remain unseen during
foundation-model training, hyperparameter tuning, and downstream fine-tuning.

\subsection{Evaluation Metrics}
We use Equal Error Rate (EER) as the main summary metric in this work. EER is the operating point at which the False Acceptance Rate (FAR) and the False Rejection Rate (FRR) are equal. The FAR denotes the proportion of unauthorized attempts incorrectly accepted by the system, whereas the FRR denotes the proportion of legitimate attempts incorrectly rejected. While EER is a widely used summary metric that facilitates comparison across studies, practical systems usually prefer very low FAR values to ensure stronger security, even at the cost of a higher FRR~\cite{fallahi2026advancing}.
Therefore, besides EER, we also report the FRR at a FAR threshold of 1\% for comparison with prior EEG-authentication studies~\cite{arias2023consumer,fallahi2023brainnet,debie2021session}, as well as at 0.1\% and 0.01\% to reflect standards-oriented biometric operating points~\cite{nist80063b2025}.

\begin{table*}[!t]
\centering
\scriptsize
\renewcommand{\arraystretch}{1.25}
\setlength{\tabcolsep}{5pt}

\definecolor{heatBest}{RGB}{198,239,206}
\definecolor{heatGood}{RGB}{226,239,218}
\definecolor{heatMid}{RGB}{255,242,204}
\definecolor{heatBad}{RGB}{252,229,205}
\definecolor{heatWorst}{RGB}{244,204,204}

\newcommand{\cellbest}[2]{\cellcolor{heatBest}#1 $\pm$ {\scriptsize #2}}
\newcommand{\cellgood}[2]{\cellcolor{heatGood}#1 $\pm$ {\scriptsize #2}}
\newcommand{\cellmid}[2]{\cellcolor{heatMid}#1 $\pm$ {\scriptsize #2}}
\newcommand{\cellbad}[2]{\cellcolor{heatBad}#1 $\pm$ {\scriptsize #2}}
\newcommand{\cellworst}[2]{\cellcolor{heatWorst}#1 $\pm$ {\scriptsize #2}}

\caption{Heatmap view of the verification results on PEERS and FRC-EEG. Greener cells indicate lower EER within each dataset row and therefore better performance for that condition.}
\label{Nasdaq}
\begin{tabular}{|c|c|c|c|c|c|c|c|}
\hline
\multicolumn{2}{|c|}{Datasets} & \textbf{Baseline} & \textbf{NS-Scratch} & \textbf{NS-ZeroShot} & \textbf{NS-FullFT} & \textbf{NS-LastFT} & \textbf{NS-PEFT} \\
\hline
\multirow{4}{*}{{PEERS}}
& All data
& \cellmid{10.25}{8.04}
& \cellgood{9.28}{6.41}
& \cellworst{16.79}{10.27}
& \cellbest{7.71}{5.81}
& \cellgood{8.23}{5.95}
& \cellbad{10.90}{7.12} \\
\cline{2-8}
& Geodesic
& \cellmid{9.18}{6.13}
& \cellbad{11.40}{5.70}
& \cellworst{13.50}{7.26}
& \cellgood{8.74}{6.20}
& \cellgood{9.04}{6.44}
& \cellbest{8.52}{5.63} \\
\cline{2-8}
& HydroCel
& \cellmid{9.88}{8.76}
& \cellbad{11.96}{8.08}
& \cellworst{15.71}{11.77}
& \cellbest{8.31}{7.22}
& \cellgood{9.48}{8.08}
& \cellmid{10.13}{9.22} \\
\cline{2-8}
& BioSemi
& \cellworst{19.56}{9.51}
& \cellbad{19.23}{7.51}
& \cellmid{17.14}{8.26}
& \cellbest{11.50}{5.80}
& \cellgood{11.51}{6.14}
& \cellgood{12.39}{5.96} \\
\hline
FRC-EEG & Emotiv Epoc X
& \cellmid{18.18}{14.12}
& \cellworst{24.92}{6.52}
& \cellbad{21.05}{5.82}
& \cellbest{14.75}{4.83}
& \cellgood{14.93}{5.76}
& \cellmid{18.76}{5.41} \\
\hline
\end{tabular}
\end{table*}

Authentication metrics are computed from verification scores
obtained by comparing enrollment templates with verification
samples. In representation-based systems such as NeuroShield,
each score is the similarity between an enrollment representation
and a verification representation in the learned embedding space.
To reflect practical deployment, enrollment should precede
verification in time. We therefore use the first session of each
subject to construct the enrollment template and use the remaining
sessions for verification. By evaluating enrollment and verification
samples from different sessions, this protocol reduces the risk of
overestimating performance due to short-term physiological states or
session-specific artifacts rather than stable subject identity.
Genuine trials compare a verification sample with the enrollment template of the same
subject, whereas impostor trials compare it with enrollment
templates from other subjects.

\subsection{Downstream Transfer Settings}

To test whether NeuroShield can serve as a reusable foundation model for EEG authentication, we evaluate downstream target datasets under settings that systematically disentangle the effects of architecture, heterogeneous pretraining, and target-specific adaptation. This comparison is necessary because an improvement on a downstream dataset could result from the flexible architecture itself, from knowledge transferred through pretraining, or from fine-tuning on the target dataset.

First, NS-Scratch trains the NeuroShield architecture from random initialization on the downstream target training split. This setting tests whether the architecture itself is competitive without transferred knowledge. Second, NS-ZeroShot applies the pretrained NeuroShield directly to the unseen target dataset without any parameter updates. This setting tests whether heterogeneous pretraining alone transfers identity-discriminative structure to a new authentication setting. Third, the fine-tuning variants adapt the pretrained model on the target training split, testing whether transferred knowledge can be converted into stronger downstream verification performance.

The fine-tuning variants differ in how much of the pretrained model is updated. NS-FullFT updates all model parameters and therefore represents the strongest form of downstream adaptation. NS-LastFT freezes most parameters and updates only the last two channel-transformer blocks, the last two temporal-transformer blocks, the learnable channel-position module, and the channel and temporal classification tokens, yielding a partial adaptation setting. NS-PEFT freezes the transformer blocks and updates only the learnable channel-position module together with the channel and temporal classification tokens, yielding a lightweight adaptation setting.

\subsection{Benchmark Baseline}
For these comparisons, we use Fallahi et al.~\cite{fallahi2026advancing} as the primary baseline. Published in 2026, this work provides a recent and comprehensive large-scale evaluation of multi-session EEG authentication and achieves state-of-the-art performance compared with previously reported methods in this setting. It is therefore a strong representative of conventional fixed-input EEG authentication pipelines that achieve high performance in dataset-specific settings, but require training a new model for each target dataset or input configuration. The study also includes extensive hyperparameter tuning across deep-learning architectures and metric-learning objectives, identifying a ResNet1D encoder trained with supervised contrastive loss (SupConLoss)~\cite{khosla2020supervised} as the best-performing configuration. We therefore reproduce this configuration as our benchmark baseline. Following the original protocol, we train and evaluate the baseline on PEERS using same subject-level train/validation/test splits. Under this aligned setup, our reproduced baseline closely matches the reported PEERS all-data result (10.25\% vs.\ 10.28\% EER).

\section{Results and Discussion}

NeuroShield is designed as a reusable foundation model for EEG authentication and is instantiated through a flexible-input architecture that supports variation in headset, channel layout, and signal duration. We develop NeuroShield using the architecture described in Section~\ref{sec:design} and the datasets and experiment protocols defined in Section~\ref{sec:methodology}, then evaluate how well the resulting model aligns with these design goals. Before the main evaluation, we determine the final architecture and implementation configuration through hyperparameter tuning, summarized in Section~\ref{sec:hpo}; all subsequent experiments use this fixed best configuration. We then evaluate the effectiveness of NeuroShield by comparing downstream transfer against training from scratch and the reproduced state-of-the-art baseline, testing whether heterogeneous pretraining improves authentication in unseen settings. Next, we test input flexibility by varying signal duration, channel layout, and available verification-time channels. Finally, we analyze channel-level spatial behavior and training-data composition to explain which model and data properties support reusable subject-discriminative representations.

\subsection{Transfer and Input Flexibility}
\label{sec:eval-main-claims}

\subsubsection{Transfer to Unseen Authentication Settings} We first test whether heterogeneous pretraining provides reusable identity information for downstream EEG authentication. If NeuroShield learns transferable subject-discriminative structure, NS-ZeroShot should yield EERs clearly below the chance-level reference on unseen datasets, while downstream fine-tuning should further improve performance over both training NeuroShield from scratch and the reproduced state-of-the-art baseline.

The downstream-transfer results summarized in Table~\ref{Nasdaq} show four main patterns. First, training the NeuroShield architecture from random initialization (NS-Scratch) indicates that the architecture is competitive with the reproduced baseline even without pretraining, but not consistently superior to it. It improves the PEERS all-data result from 10.25\% to 9.28\% EER and slightly improves the BioSemi split from 19.56\% to 19.23\% EER, but remains weaker than the baseline on Geodesic, HydroCel, and FRC-EEG. Second, directly applying the pretrained model without target-dataset updates (NS-ZeroShot) shows that heterogeneous pretraining transfers identity-discriminative structure to unseen target datasets. Across all evaluated settings, its EER remains clearly below the 50\% chance-level reference, ranging from 13.50\% to 21.05\%. However, zero-shot performance is generally weaker than the reproduced baseline and NS-Scratch. This indicates that heterogeneous pretraining transfers useful identity information to unseen datasets, but that target-specific adaptation is still needed for competitive downstream performance. Third, among the fine-tuning strategies, full-model fine-tuning (NS-FullFT) generally performs best, while partial fine-tuning of the last transformer layers (NS-LastFT) remains close in several settings and parameter-efficient fine-tuning of only the lightweight adaptation modules (NS-PEFT) is less consistent.

This indicates that adapting the full pretrained representation is more effective than updating only a subset of parameters. Fourth, after downstream fine-tuning, NeuroShield consistently outperforms the reproduced state-of-the-art baseline, most notably NeuroShield reduces the PEERS all-data EER from 10.25\% to 7.71\%, corresponding to a 2.54 percentage-point gain over the state-of-the-art. The result is measured on the largest and most comprehensive multi-session EEG authentication dataset in the literature, with recordings collected over more than five years. Since NS-FullFT also improves over NS-Scratch, these results indicate that the strongest performance is not explained by the architecture alone, but by the combination of flexible architecture, heterogeneous pretraining, and target-specific fine-tuning.

\begin{figure*}[!t]
    \centering
    \begin{subfigure}[b]{0.65\textwidth}
        \centering
        \includegraphics[width=\linewidth]{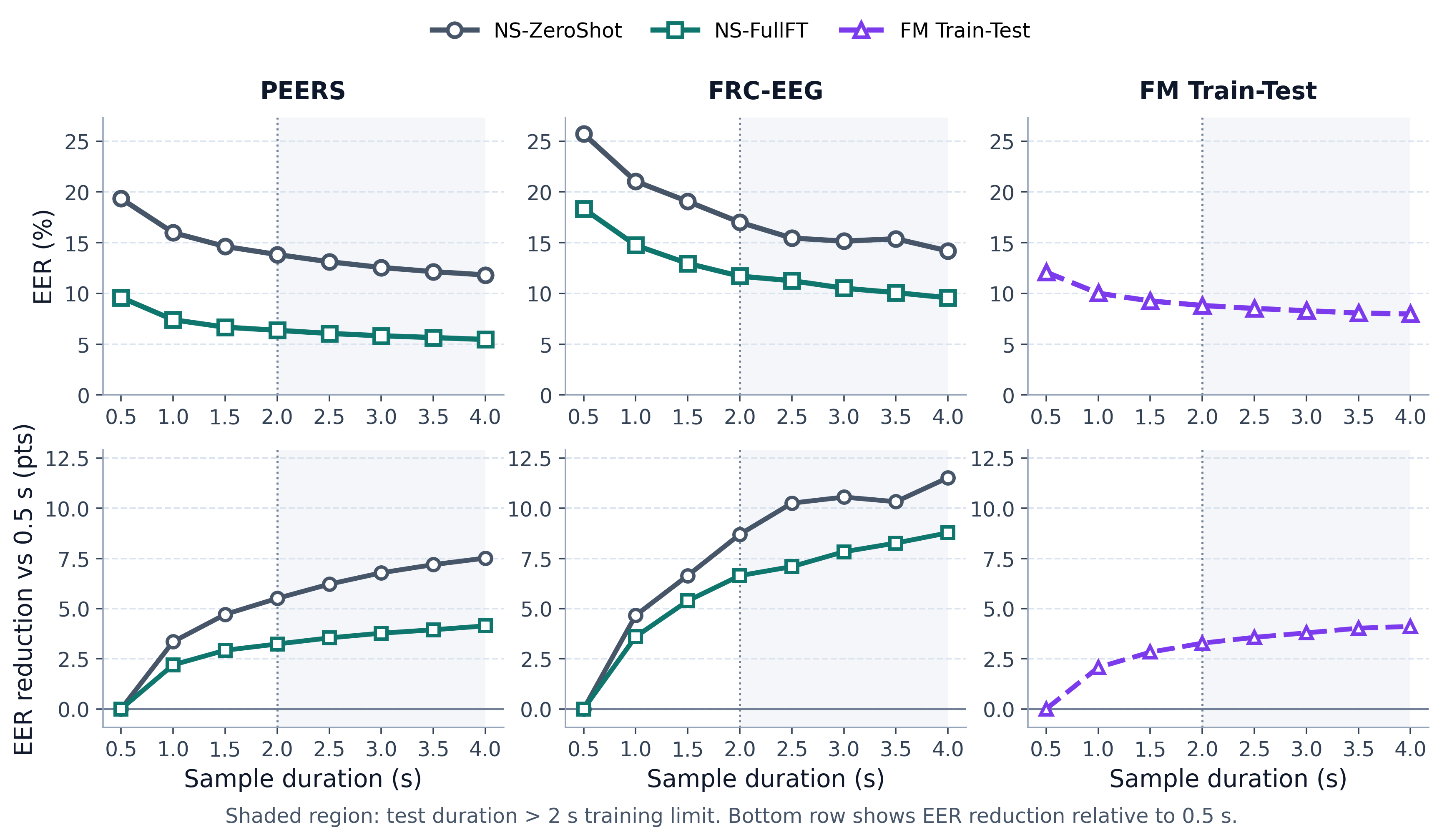}
        \caption{Temporal flexibility across input durations.}
        \label{fig:variable_length}
    \end{subfigure}
    \hfill
    \begin{subfigure}[b]{0.33\textwidth}
        \centering
        \raisebox{15mm}{\includegraphics[width=\linewidth]{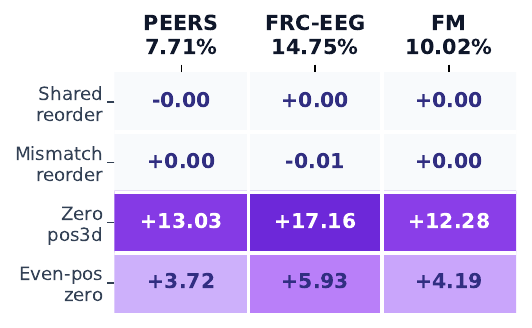}}
        \caption{Channel-layout perturbations.}
        \label{fig:channel_flexibility}
    \end{subfigure}

    \caption{Input-flexibility analysis.
    (a) Temporal flexibility across PEERS, FRC-EEG, and FM Train-Test. The top row reports absolute EER as a function of input duration, while the bottom row reports EER reduction relative to the \(0.5\)-second condition. NS-ZeroShot and NS-FullFT are shown for the downstream datasets, and FM Train-Test summarizes the foundation-model data split. The shaded region marks test durations beyond the \(2.0\)-second training-time limit.
    (b) Channel-layout perturbation results. Cells report EER change relative to the corresponding baseline. Shared reorder applies the same channel permutation to enrollment and verification, while mismatch reorder uses different permutations. Zero pos3d removes all 3D channel coordinates, and even-pos zero removes coordinates only for channels whose labels end in an even digit.}
    \label{fig:input_flexibility}
\end{figure*}

\begin{figure*}[!b]
    \centering
    \includegraphics[width=0.80\textwidth]{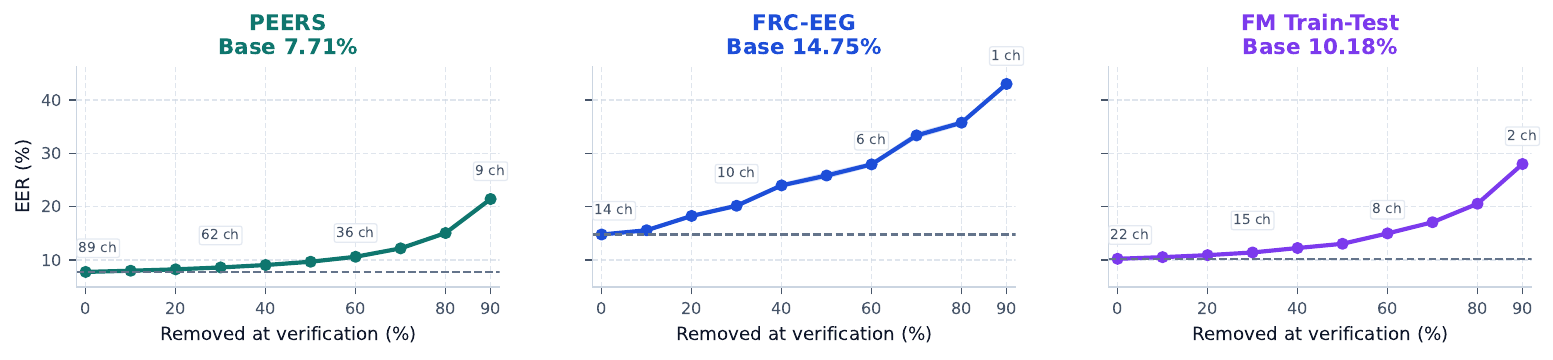}
    \caption{Robustness to random missing verification channels across PEERS, FRC-EEG, and FM Train-Test. Enrollment uses all channels, while channels are randomly removed only from verification inputs.}
    \label{fig:random_missing_multidataset}
\end{figure*}

\subsubsection{Temporal Flexibility Across Signal Durations}
\label{sec:eval-temporal}
We next test the extent to which NeuroShield can benefit from longer EEG input segments, including durations longer than those observed during training. If the temporal-flexibility design is effective, increasing the test-time duration should reduce EER as more temporal evidence becomes available, and this improvement should continue beyond the 2.0-second training-time limit.
Figure~\ref{fig:variable_length}
summarizes EER from \(0.5\) to \(4.0\) seconds for
NS-ZeroShot and NS-FullFT on PEERS and FRC-EEG, together
with FM Train-Test on the foundation-model data split,
while the bottom row reports EER reduction relative
to the \(0.5\)-second condition. During training,
NeuroShield only saw samples up to \(2.0\) seconds,
so the shaded region highlights evaluation beyond that
limit, where we expect the temporal transformer to
generalize to longer inputs via fixed ALiBi biases.
Across all reported settings, EER decreases as more
temporal context becomes available, and the gain continues
beyond \(2.0\) seconds rather than saturating at the
training-time limit. For example, under NS-FullFT, PEERS
improves from \(9.60\%\) at \(0.5\) seconds to
\(5.47\%\) at \(4.0\) seconds, FRC-EEG improves from
\(18.35\%\) to \(9.57\%\), and FM Train-Test improves
from \(12.09\%\) to \(7.98\%\). Taken together, these
results support the claim that NeuroShield remains effective
across variable input durations and can generalize to
longer temporal windows than those seen during training.
The 5.47\% EER is the lowest PEERS result at 4 seconds,
improving over the prior state-of-the-art
result of 6.22\%~\cite{fallahi2026advancing}.

\begin{figure*}[!t]
    \centering
    \begin{subfigure}[b]{0.64\textwidth}
        \centering
        \includegraphics[width=\linewidth]{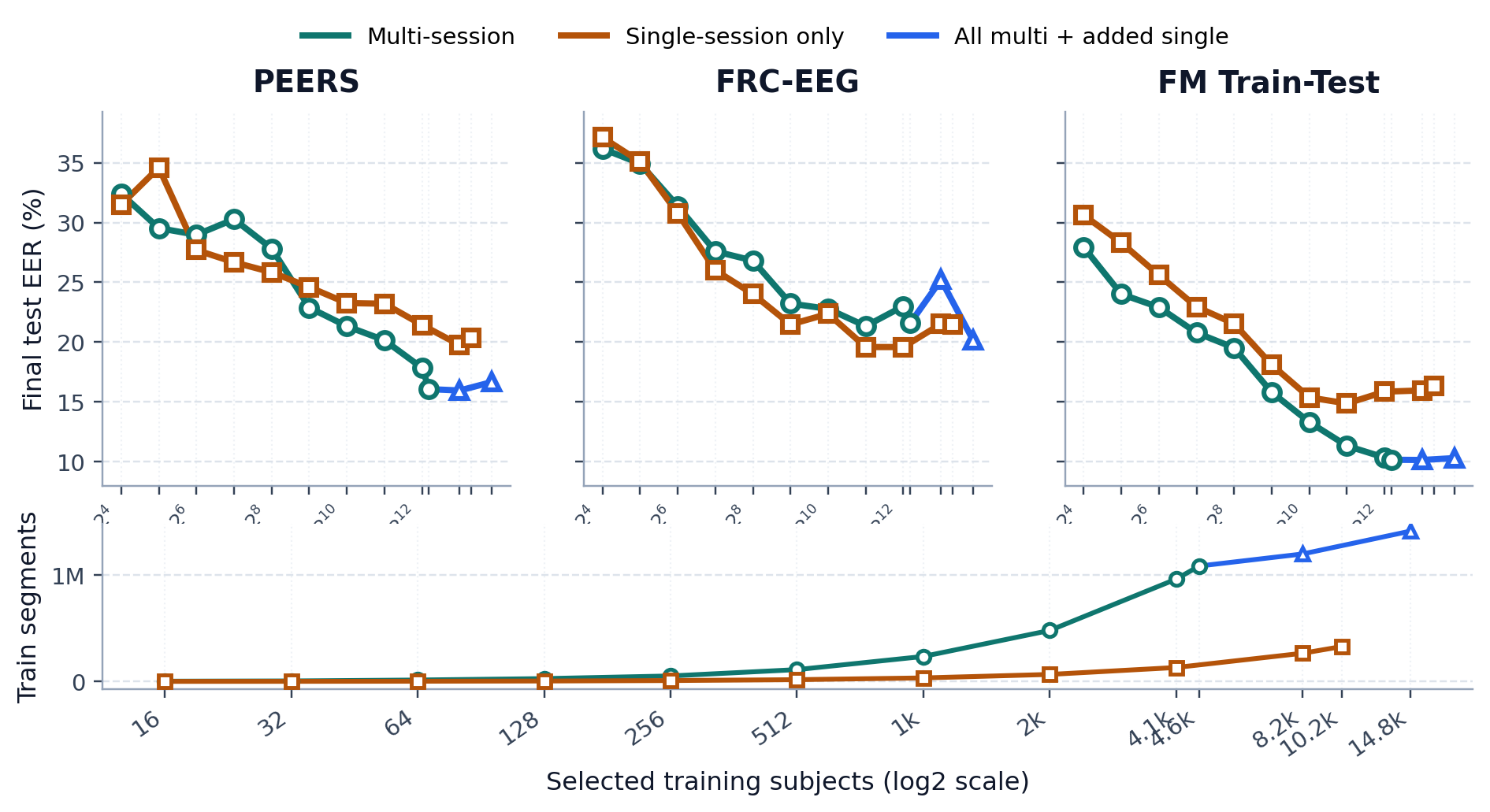}
        \caption{Effect of training-data composition.}
        \label{fig:session_scaling}
    \end{subfigure}
    \hfill
    \begin{subfigure}[b]{0.31\textwidth}
        \centering
        \raisebox{6mm}{\includegraphics[width=\linewidth]{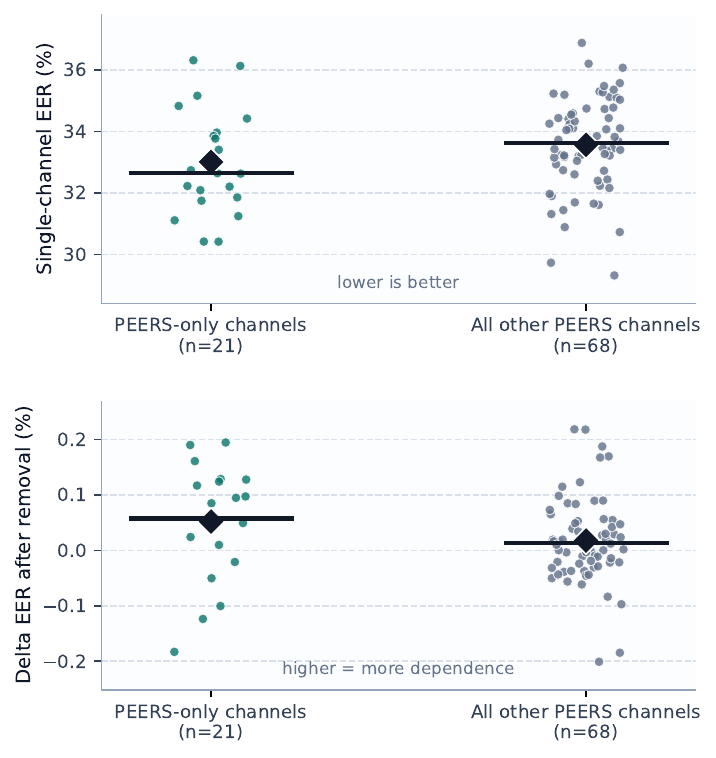}}
        \caption{Seen and unseen PEERS channels.}
        \label{fig:peers_fm_absent_channels}
    \end{subfigure}

    \caption{Training-data and channel-generalization analysis.
    (a) Effect of training-data composition as the number of selected training subjects increases. Green shows multi-session-only training, orange shows single-session-only training, and blue shows the extension in which all available multi-session subjects are first used and then additional single-session-only subjects are added. The top row reports final test EER and the bottom row reports the corresponding number of training segments.
    (b) Channel-level comparison between 21 PEERS-only channels, whose channel names were not present in the training datasets, and 68 PEERS channels whose names were present during training. The left panel reports single-channel-only EER, and the right panel reports leave-one-out delta EER versus the full baseline.}
    \label{fig:training_channel_analysis}
\end{figure*}

\subsubsection{Channel-Layout Flexibility and Position Encoding}
\label{sec:eval-spatial}
The spatial flexibility claim is already supported
partly by Table~\ref{Nasdaq},
where the NS-ZeroShot results show that the
NeuroShield foundation model can still operate on
unseen downstream datasets even when the input
channel set differs from the source training
data. However, that evidence alone does not
show whether the model relies primarily on meaningful
3D channel geometry or simply on the order
in which channels are presented, as is commonly
used in other EEG authentication models. We therefore
evaluate this distinction more directly in
Figure~\ref{fig:channel_flexibility}, which reports the
change in EER relative to each dataset's own
baseline under four controlled perturbations. Here,
channel permutation refers to reordering the input
channel-token sequence without changing the EEG
signals or their associated channel coordinates.
same global channel permutation is applied to both
enrollment and verification, the effect on performance
is negligible across all three datasets, and even
using different permutations at enrollment and verification
causes only minor degradation. This indicates that
NeuroShield is not strongly tied to a canonical
channel ordering. In contrast, removing spatial
coordinates causes a large performance drop, while
zeroing the coordinate information for only part
of the montage produces a smaller but still
consistent degradation. Taken together, these results
show that NeuroShield achieves effective spatial flexibility
by using spatial channel information rather than
fixed channel order, which helps it generalize
across new datasets and recording hardware.

\subsubsection{Robustness to Missing Channels}
\label{sec:missing-channels}

A practical advantage of NeuroShield’s spatial flexibility is that verification does not require exactly the same complete channel set used during enrollment. This is important when some verification channels are unavailable because of sensor noise, poor contact, or hardware defects. To test this setting, we evaluate random channel removal at verification across PEERS, FRC-EEG, and FM Train-Test, while enrollment uses the full channel set. Figure~\ref{fig:random_missing_multidataset} reports the resulting EER as the verification-time channel-removal rate increases. EER increases gradually as more verification channels are removed, and the degradation rate depends strongly on montage density. Nevertheless, all three settings remain close to baseline up to about 10\% missing channels. These results show that NeuroShield can tolerate a small number of noisy or missing channels without requiring the entire EEG segment to be discarded.

\subsubsection{Robustness to unseen channel through training}
Channel flexibility is a central goal of NeuroShield. Although our cross-headset results show that the model can operate across different layouts, many electrodes are shared across headsets, so those experiments do not fully test generalization to entirely unseen channels. To isolate that question, we identified 21 PEERS channels that were absent from the foundation-model training datasets and compared them with the remaining 68 PEERS channels that were represented during pretraining. We expected the geometry-aware 3D positional encoder to generalize to these unseen electrodes, so the two groups should not differ substantially.

We tested this using NeuroShield without fine-tuning. For each channel, we measured single-channel-only EER to capture standalone utility and leave-one-out delta EER to measure how much the full model depends on that channel, then compared the two channel groups using two-sided Monte Carlo permutation test~\cite{ernst2004permutation}. The unseen PEERS-only channels showed slightly lower single-channel EER than the seen-channel group (33.01\% vs 33.57\%; p = 0.153) and slightly larger leave-one-out impact (0.052 vs 0.018 delta EER; p = 0.0965), but neither difference was statistically significant. Overall, these results suggest that NeuroShield is not limited to training-time channel identities and can extend its learned representation to previously unseen electrodes.

\subsection{Ablation and Design Analysis}

\begin{table*}[t]
\centering
\caption{Summary of hyperparameter tuning outcomes for the final
NeuroShield model. For each component, we report the search values,
the selected value, how often that selected value appears among
the top 100 and top 1000 completed trials, its relative enrichment
in the top 100 compared with the top 1000, and its PED-ANOVA
importance.}
\label{tab:hpo_summary}
\footnotesize
\setlength{\tabcolsep}{2pt}
\renewcommand{\arraystretch}{1.10}
\begin{tabular*}{\textwidth}{@{\extracolsep{\fill}}
>{\raggedright\arraybackslash}p{2.95cm}
>{\raggedright\arraybackslash}p{4.00cm}
>{\centering\arraybackslash}p{1.25cm}
>{\centering\arraybackslash}p{1.00cm}
>{\centering\arraybackslash}p{1.15cm}
>{\centering\arraybackslash}p{1.30cm}
>{\centering\arraybackslash}p{1.15cm}
@{}}
\toprule
\multirow{2}{*}{Component} &
\multirow{2}{*}{Search space} &
\multirow{2}{*}{Selected} &
\multicolumn{2}{c}{\shortstack{Sel.-value\\freq.}} &
\multirow{2}{*}{\shortstack{Overrep.\\ratio}} &
\multirow{2}{*}{\shortstack{PED\\(\%)}} \\
\cmidrule(lr){4-5}
& & & Top 100 & Top 1000 & & \\
\midrule
Patch embedder &
linear / mlp\_patch / cnn\_simple / cnn\_multiscale &
mlp\_patch & 34/100 & 247/1000 & 1.38$\times$ & 1.67 \\

Spatial coordinate scale &
none / 100 &
100 & 93/100 & 502/1000 & 1.85$\times$ & 30.98 \\

Channel normalization &
none / zscore / rms &
none & 100/100 & 353/1000 & 2.83$\times$ & 37.23 \\

Embedding dimension &
64 / 128 &
128 & 60/100 & 466/1000 & 1.29$\times$ & 1.68 \\

Attention heads &
4 / 8 &
8 & 55/100 & 475/1000 & 1.16$\times$ & 0.42 \\

Channel depth &
2 / 4 / 6 &
6 & 38/100 & 317/1000 & 1.20$\times$ & 2.65 \\

Temporal depth &
2 / 4 &
4 & 54/100 & 517/1000 & 1.04$\times$ & 0.27 \\

Embedding norm &
none / l2 &
l2 & 54/100 & 508/1000 & 1.06$\times$ & 0.27 \\

Patch length &
50 / 100 &
50 & 48/100 & 465/1000 & 1.03$\times$ & 0.07 \\

Mask ratio &
0 / 0.1 / 0.2 &
0.2 & 26/100 & 327/1000 & 0.80$\times$ & 2.38 \\

Learning rate &
1e-5 / 1e-4 / 1e-3 &
1e-4 & 29/100 & 329/1000 & 0.88$\times$ & 22.24 \\

Weight decay &
0 / 1e-5 &
1e-5 & 47/100 & 478/1000 & 0.98$\times$ & 0.15 \\
\bottomrule
\end{tabular*}
\end{table*}

\subsubsection{Architecture Selection and Hyperparameter Effects}
\label{sec:hpo}
We next analyze the outcome of hyperparameter tuning
to understand which design choices most consistently
supported low validation EER in the final NeuroShield
model, using the best-trial configuration, top-trial
frequencies, and PED-ANOVA~\cite{watanabe2023ped} importance analysis.
Table~\ref{tab:hpo_summary} summarizes these outcomes
over the completed Optuna study. The overrepresentation
ratio and PED-ANOVA serve complementary purposes.
The overrepresentation ratio compares how often the
selected value appears in the top 100 trials
relative to its frequency across all completed
trials. PED-ANOVA also focuses on the top-performing
trials, but instead of counting only how often
the selected value appears, it compares the
full hyperparameter distribution of that top
subset against the completed study.

Across 1000 completed trials, the best trial
in the tuning study used an MLP-based patch
embedder, explicit spatial coordinate scaling, no
additional channel normalization, 128-dimensional embeddings,
8 attention heads, channel depth 6, temporal
depth 4, L2 embedding normalization, patch length
50, mask ratio 0.2, learning rate \(10^{-4}\),
and weight decay \(10^{-5}\). The search space
was intentionally bounded by hardware feasibility,
so that all sampled configurations remained trainable
under the worst-case resource usage considered in
the study. Several best-trial values are also
clearly overrepresented among the strongest runs.
In particular, no channel normalization appears in
all top 100 trials, and explicit spatial
coordinate scaling appears in 93 of the top
100 trials. The MLP-based patch embedder is
likewise more common in the top 100 than
across all trials, although its overrepresentation
is more moderate than that of the normalization
and coordinate-scaling choices. Under PED-ANOVA,
the largest importance scores are assigned to
channel normalization (37.23\%), spatial coordinate
scaling (30.98\%), and learning rate (22.24\%),
while the remaining hyperparameters together account
for the rest of the total importance.

Taken together, these results suggest that the
final NeuroShield configuration was not the outcome
of a single favorable trial, but emerged from
consistent patterns across the tuning study.
The strongest differences in validation EER were
associated primarily with channel normalization,
coordinate scaling, and learning rate, while
the strongest configurations more often combined
the MLP-based patch embedder with the selected
higher-capacity architectural settings.

\subsubsection{Effect of Pretraining Data Composition}
\label{sec:training-data}

To understand what part of heterogeneous pretraining supports transfer, we separate the effect of dataset scale from the effect of session diversity. This matters because EEG authentication depends on identity features that persist from enrollment to later verification, not features specific to one recording session. We therefore expect multi-session training data to support stronger transfer than single-session data when the number of training subjects increases.

Figure~\ref{fig:session_scaling} shows that multi-session data is generally more effective than single-session data as the number of training subjects increases. This pattern is clearest on PEERS and FM Train-Test, where multi-session data consistently reaches lower EER across most of the range. FRC-EEG does not show the same consistency, which likely reflects both its smaller dataset size and the more limited temporal separation between enrollment and verification samples, since the dataset records the first two sessions on the same day with a 1--2 hour interval and the remaining sessions after 2--3 days. Importantly, once the full multi-session pool is used, adding single-session-only subjects substantially increases the amount of training data but does not reliably improve EER. This suggests that the benefit of additional training data depends strongly on session diversity, and that repeated recordings per subject are more valuable than simply adding more single-session samples.

\begin{figure*}[!t]
    \centering
    \includegraphics[width=0.80\textwidth]{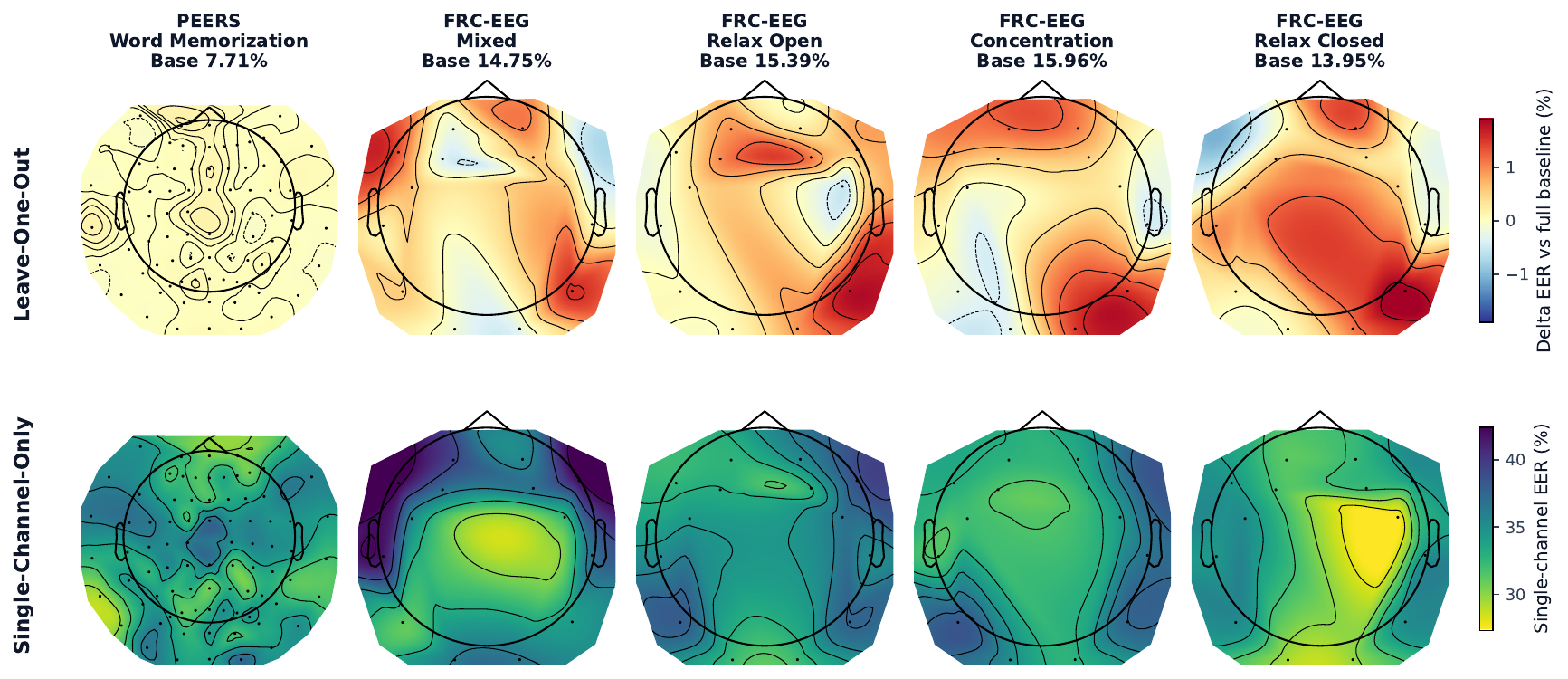}
    \caption{Channel-level spatial analysis across PEERS and FRC-EEG. The top row shows leave-one-channel-out results, where one channel is removed at a time and the scalp color reports the resulting change in EER relative to the full-channel baseline. The bottom row shows single-channel-only results, where only one channel is retained and the scalp color reports the resulting EER for that channel alone. The FRC-EEG panels include both the mixed benchmark setting and task-specific views. Channel aliases that share the same scalp position are merged only for visualization.}
    \label{fig:channel_topomap_compact}
\end{figure*}

\subsubsection{Spatial Contribution of EEG Channels}

EEG responses can exhibit different spatial patterns across tasks and stimuli. In EEG authentication, this suggests that the channels contributing most to verification may also depend on the acquisition task. Identifying such task-dependent spatial contributions is relevant for deployment because it can inform both headset selection for a given authentication task and task selection for a given headset layout. NeuroShield enables this analysis because it can operate with different channel subsets at evaluation time, allowing us to remove or retain channels without retraining the network and making the resulting performance changes directly comparable across channels.

Figure~\ref{fig:channel_topomap_compact} suggests that channel-level authentication error is partly related to the task performed during EEG recording. In PEERS, where we use a word-memorization task, the dense montage makes each leave-one-channel-out effect small, but sensitivity appears mainly around frontal regions, with smaller left-temporal and parietal contributions. This is compatible with verbal memory tasks, where successful memory encoding has been associated with frontal theta activity~\cite{long2014subsequent}. In FRC-EEG, the task-specific maps show a different pattern, with the most visible contribution appearing in posterior regions, especially during eyes-closed relaxation, and smaller frontal contributions across the tasks. This is compatible with EEG literature showing that posterior alpha is prominent during resting tasks~\cite{tenke2015posterior}.

Together, these results suggest that the regions contributing to authentication error are not fixed across datasets, but are modulated by the cognitive or resting task used during acquisition. This also implies that task design and headset layout should be considered jointly, since a task may be more suitable for authentication when the available channels cover the regions where subject-discriminative information is expressed. At the same time, the single-channel-only maps show that no individual channel is sufficient for robust authentication, indicating that NeuroShield relies on distributed, task-dependent spatial information.

\subsection{Comparison with Prior EEG Works}

\label{sec:comparison}

To contextualize NeuroShield relative to prior EEG biometric systems, Table~\ref{tab:sota_comparison} compares it with recent studies. We organize the comparison around three relevant modeling directions: CNN-based authentication pipelines, which represent strong EEG authentication baselines in comparable evaluation settings~\cite{fallahi2026advancing}; transformer-based EEG biometric models, which are architecturally closest to NeuroShield~\cite{li2025enhancing}; and general EEG foundation models, which assess the extent to which reusable EEG representations can transfer to authentication~\cite{jiang2024labram}.

\begin{table*}[t]
\centering
\scriptsize
\setlength{\tabcolsep}{3pt}
\renewcommand{\arraystretch}{1.12}
\begin{threeparttable}
\caption{Representative EEG biometric systems and transfer baselines.}
\label{tab:sota_comparison}
\begin{tabular*}{\textwidth}{@{\extracolsep{\fill}}
>{\raggedright\arraybackslash}m{1.75cm}
>{\centering\arraybackslash}m{0.52cm}
>{\centering\arraybackslash}m{0.72cm}
>{\centering\arraybackslash}m{0.90cm}
>{\centering\arraybackslash}m{0.72cm}
>{\centering\arraybackslash}m{0.72cm}
>{\raggedright\arraybackslash}p{2.05cm}
>{\raggedright\arraybackslash}p{1.35cm}
>{\raggedright\arraybackslash}p{1.40cm}
>{\centering\arraybackslash}p{0.90cm}
>{\centering\arraybackslash}p{0.78cm}
>{\centering\arraybackslash}p{0.90cm}
>{\raggedright\arraybackslash}p{2.70cm}
@{}}
\toprule
Paper &
Year &
Open &
\shortstack{Pretr.\\Ckpt.} &
\shortstack{Temp.\\Flex.} &
\shortstack{Spatial\\Flex.} &
Train Data &
Eval Data &
\shortstack{Subj. /\\Tot. sess.} &
\shortstack{Device\\grade} &
Dur. &
Metric &
\shortstack{Result\\(\%)} \\
\midrule
\multirow{2}{*}{Li et al.~\cite{li2025enhancing}} &
\multirow{2}{*}{2025} &
\multirow{2}{*}{$\checkmark$} &
\multirow{2}{*}{$\checkmark$} &
\multirow{2}{*}{$\times$} &
\multirow{2}{*}{$\times$} &
\multirow{2}{*}{\shortstack[l]{BCI IV 2a~\cite{tangermann2012review};\\PhysioNetMI~\cite{schalk2004bci2000}}} &
BCI IV 2a &
9 / 18 &
Medical &
3 s &
\shortstack{Acc.} &
97.65 \\
&
&
&
&
&
&
&
PhysioNetMI &
109 / 109 &
Medical &
4 s &
\shortstack{Acc.} &
91.81 \\
\midrule
Maiorana~\cite{maiorana2021taskindependent} &
2021 &
$\times$ &
$\times$ &
$\times$ &
$\triangle$ &
Private dataset &
\mbox{Same dataset} &
45 / 225 &
Medical &
5 s &
EER &
4.8--10.7 \\
\midrule
Fallahi et al.~\cite{fallahi2026advancing} &
2026 &
$\checkmark$ &
$\checkmark$ &
$\times$ &
$\times$ &
PEERS~\cite{kahana2024penn} &
PEERS &
\shortstack[c]{345/6007} &
Medical &
1 s &
EER &
10.28 \\
\midrule
\multirow{2}{*}{Jiang et al.~\cite{jiang2024labram}\tnote{a}} &
\multirow{2}{*}{2024} &
\multirow{2}{*}{$\checkmark$} &
\multirow{2}{*}{$\checkmark$} &
\multirow{2}{*}{$\checkmark$} &
\multirow{2}{*}{$\triangle$} &
\multirow{2}{*}{16 ds.; 1.5K subj.\tnote{b}} &
PEERS~\cite{kahana2024penn} &
345 / 6007 &
Medical &
1 s &
EER &
ZS/FT: 44.32/ 10.27 \\
&
&
&
&
&
&
&
FRC.~\cite{albasri2019eeg} &
30 / 120 &
Consumer &
1 s &
EER &
ZS/FT: 47.16 / 31.25 \\
\midrule
\multirow{2}{*}{\shortstack[l]{\textbf{This work}\\\textbf{(NeuroShield)}}} &
\multirow{2}{*}{2026} &
\multirow{2}{*}{$\checkmark$} &
\multirow{2}{*}{$\checkmark$} &
\multirow{2}{*}{$\checkmark$} &
\multirow{2}{*}{$\checkmark$} &
\multirow{2}{*}{\shortstack[l]{3 ds.; 15.7K subj.\tnote{c}}} &
PEERS~\cite{kahana2024penn} &
345 / 6007 &
Medical &
\shortstack{1 s} &
\shortstack{EER} &
ZS/FT: 16.79 / 7.71 \\
&
&
&
&
&
&
&
FRC.~\cite{albasri2019eeg} &
30 / 120 &
Consumer &
\shortstack{1 s} &
\shortstack{EER} &
ZS/FT: 21.05 / 14.75 \\
\bottomrule
\end{tabular*}
\begin{tablenotes}[flushleft]
\scriptsize
\item Symbols: $\checkmark$ = yes, $\times$ = no, $\triangle$ = partial.
\item[a] We report our zero-shot and fine-tuned results on our evaluation datasets using the LaBraM foundation model.
\item[b] The Jiang et al. pretraining summary is taken from Jiang et al.~\cite[App.~D]{jiang2024labram}.
\item[c] NeuroShield training datasets are TUH-EEG~\cite{harati2014tuh}, DVS-rsEEG~\cite{getzmann2024resting}, and PREDICT~\cite{seminowicz2020novel}.
\end{tablenotes}
\end{threeparttable}
\end{table*}

Among mainstream EEG authentication studies that rely
on CNN-based pipelines, we compare primarily with
Maiorana et al.~\cite{maiorana2021taskindependent},
one of the earlier works to explicitly investigate
multi-session EEG verification. That method learns
channel-specific deep representations and combines
them with one-class SVM scoring, which gives a
limited form of channel flexibility only if a
pretrained model is available for each channel
separately. The study evaluates 15 held-out subjects
after training and validation on the remaining 30
subjects and reports EERs ranging from 4.8\% to
10.7\% across different task settings. However, two
characteristics make direct reproduction and practical
reuse difficult: the method does not model
cross-channel interactions, and it depends on a
one-class SVM backend, which later large-scale
evaluation found to be less reliable than
metric-learning alternatives~\cite{fallahi2026advancing}. Moreover, in the absence of publicly available datasets, evaluation protocols, and open-source code, and given the challenges discussed above, we were unable to reproduce the reported results, while our implementation yielded substantially higher EERs. We therefore also compare with Fallahi
et al.~\cite{fallahi2026advancing}, which is the most
recent multi-session EEG authentication study with
publicly available code and data, reserves 100
subjects for evaluation, and serves as our primary
reproduced baseline. When evaluated within our PEERS pipeline, this reproduced architecture closely matches the originally reported performance (10.25\% vs.\ 10.28\% EER), indicating that our implementation follows an aligned evaluation protocol. Under the same setting, fine-tuned NeuroShield further reduces EER to 7.71\% while additionally supporting variation in channel layout and input duration.

Transformer-based EEG biometrics have gained attention
in recent years, but most studies still focus on
closed-set identification rather than verification-style
authentication, including ETST by Du
et al.~\cite{du2022etst}, AITST by Cai
et al.~\cite{cai2023aitst}, and CLT by Shao
et al.~\cite{shao2026clt}. In these works, the
transformer is used mainly as a stronger sequence
model to improve recognition performance within a
fixed dataset setting, rather than to support
reusable flexibility across datasets, channel layouts,
or input durations. We therefore select Li
et al.~\cite{li2025enhancing} as the most relevant
transformer-based comparison because it is explicitly
framed as EEG authentication. However, its
implementation and evaluation remain closer to
identification than to verification-style authentication.
The experiments are configured separately for each
dataset, fixed input shapes are assumed in the
released code,\footnote{\url{github.com/snow1/transformer}}
and results are reported as mean accuracy rather
than verification metrics such as EER or FRR.
Thus, this line of work is relevant as a
transformer-based EEG biometric baseline, but it
does not directly address the flexible cross-dataset
verification setting targeted in this paper.

An alternative approach to EEG authentication is to leverage general EEG foundation models. However, these models are typically pretrained using self-supervised or unsupervised objectives and are designed for broad applicability across diverse EEG tasks rather than biometric verification. As a result, they are not expected to learn representations that are optimal for EEG authentication. Recently,
Meta's NeuralBench~\cite{banville2026neuralbench}
provided a useful reference point by evaluating
several EEG foundation models under a unified
evaluation pipeline. In that benchmark,
REVE~\cite{elouahidi2025reve} ranked first and
LaBraM~\cite{jiang2024labram} second, with only a
small performance gap between them. However, because
REVE does not permit biometric identification
usage,\footnote{\url{huggingface.co/brain-bzh/reve-base}}
we selected LaBraM as the comparison baseline. The results show that LaBraM performs close to chance in the zero-shot setting. On FRC-EEG, LaBraM obtains 47.16\% EER without adaptation and 31.25\% EER after fine-tuning. On PEERS, LaBraM obtains 44.32\% EER in the zero-shot setting and 10.27\% EER after fine-tuning. These results indicate that general self-supervised EEG pretraining does not directly transfer to effective EEG authentication, even though task-specific fine-tuning can substantially improve performance. This gap likely reflects both the absence of subject-identity supervision during LaBraM pretraining and architectural differences, since LaBraM was designed as a general EEG foundation model rather than an authentication-specific encoder. Overall, NeuroShield provides a more suitable representation-learning for EEG authentication.

\section{Conclusion and Future Works}
\label{sec:conclusion}

This paper presented NeuroShield, a spatially and temporally
flexible foundation model for EEG authentication. In contrast
to prior systems that are typically tied to fixed channel
layouts, fixed input durations, and dataset-specific retraining,
NeuroShield is designed to learn reusable
subject-discriminative representations from heterogeneous EEG
recordings and to transfer them across unseen downstream
datasets, devices, and recording setups.

Our results show that this formulation is effective in practice. NeuroShield produced identity-discriminative representations on unseen downstream datasets and, after fine-tuning, reduced EER by 0.5--8.0 percentage points relative to the reproduced benchmark baseline. Additional analyses showed that the model can benefit from longer test-time input windows than those seen during training and that its spatial robustness is supported by geometry-aware channel embeddings rather than fixed channel order. Together, these findings support the view that EEG authentication can be approached as a reusable representation-learning problem rather than as a sequence of isolated dataset-specific models. In this setting, NeuroShield can serve as a pretrained building block for future EEG-authentication research, providing a stronger starting point rather than requiring a new model to be trained from scratch for each dataset or device.

At the same time, the results also clarify the remaining challenges. Although NeuroShield surpasses the reproduced state-of-the-art baseline, its error rates still need to be reduced further before EEG authentication can approach the maturity of established biometric modalities such as face recognition. However, our complementary experiments suggest that this limitation may be mitigated when longer segments are available, as in continuous authentication. Future work should therefore expand pretraining to larger and more diverse EEG corpora, evaluate stronger threat models and more realistic deployment conditions, and further study transfer across consumer-grade devices and cross-protocol recording settings.

\bibliographystyle{IEEEtran}
\bibliography{ref}

@article{fidas2023review,
  title={A review of EEG-based user authentication: trends and future research directions},
  author={Fidas, Christos A and Lyras, Dimitrios},
  journal={IEEE Access},
  volume={11},
  pages={22917--22934},
  year={2023},
  publisher={IEEE},
  doi={10.1109/ACCESS.2023.3253026}
}

@article{ernst2004permutation,
  title={Permutation Methods: A Basis for Exact Inference},
  author={Michael D. Ernst},
  journal={Statistical Science},
  year={2004},
  volume={19},
  pages={676-685},
  url={https://api.semanticscholar.org/CorpusID:16746275}
}

@article{fallahi2026beyond,
  title={Beyond gaze points: augmenting eye movement with brainwave data for multimodal user authentication in extended reality},
  author={Fallahi, Matin and Arias-Cabarcos, Patricia and Strufe, Thorsten},
  journal={Complex \& Intelligent Systems},
  volume={12},
  number={1},
  pages={39},
  year={2026},
  publisher={Springer}
}

@article{jain2004introduction,
  author       = {Anil K. Jain and
                  Arun Ross and
                  Salil Prabhakar},
  title        = {An introduction to biometric recognition},
  journal      = {{IEEE} Trans. Circuits Syst. Video Technol.},
  volume       = {14},
  number       = {1},
  pages        = {4--20},
  year         = {2004},
  url          = {https://doi.org/10.1109/TCSVT.2003.818349},
  doi          = {10.1109/TCSVT.2003.818349},
  bibsource    = {dblp computer science bibliography, https://dblp.org}
}

@article{watanabe2023ped,
  title={PED-ANOVA: Efficiently quantifying hyperparameter importance in arbitrary subspaces},
  author={Watanabe, Shuhei and Bansal, Archit and Hutter, Frank},
  journal={arXiv preprint arXiv:2304.10255},
  year={2023}
}

@article{gui2019survey,
  author       = {Qiong Gui and
                  Maria V. Ruiz{-}Blondet and
                  Sarah Laszlo and
                  Zhanpeng Jin},
  title        = {A Survey on Brain Biometrics},
  journal      = {{ACM} Comput. Surv.},
  volume       = {51},
  number       = {6},
  pages        = {112:1--112:38},
  year         = {2019},
  url          = {https://doi.org/10.1145/3230632},
  doi          = {10.1145/3230632},
  bibsource    = {dblp computer science bibliography, https://dblp.org}
}

@article{marcel2007brainwaves,
  title={Person Authentication Using Brainwaves (EEG) and Maximum A Posteriori Model Adaptation},
  author={Marcel, S{\'e}bastien and del R. Mill{\'a}n, Jos{\'e}},
  journal={IEEE Transactions on Pattern Analysis and Machine Intelligence},
  volume={29},
  number={4},
  pages={743--752},
  year={2007},
  publisher={IEEE},
  doi={10.1109/TPAMI.2007.1012}
}

@article{bidgoly2020survey,
  title={A survey on methods and challenges in EEG based authentication},
  author={Bidgoly, Amir Jalaly and Bidgoly, Hamed Jalaly and Arezoumand, Zeynab},
  journal={Computers \& Security},
  volume={93},
  pages={101788},
  year={2020},
  publisher={Elsevier},
  doi={10.1016/j.cose.2020.101788}
}

@article{maiorana2021taskindependent,
  author       = {Emanuele Maiorana},
  title        = {Learning deep features for task-independent EEG-based biometric verification},
  journal      = {Pattern Recognit. Lett.},
  volume       = {143},
  pages        = {122--129},
  year         = {2021},
  url          = {https://doi.org/10.1016/j.patrec.2021.01.004},
  doi          = {10.1016/J.PATREC.2021.01.004},
  bibsource    = {dblp computer science bibliography, https://dblp.org}
}

@article{bidgoly2022universal,
  title={Towards a Universal and Privacy Preserving EEG-Based Authentication System},
  author={Bidgoly, Amir Jalaly and Bidgoly, Hamed Jalaly and Arezoumand, Zeynab},
  journal={Scientific Reports},
  volume={12},
  number={1},
  pages={2531},
  year={2022},
  publisher={Nature Publishing Group UK London},
  doi={10.1038/s41598-022-06527-7}
}

@article{zhao2026brainwave,
  title={A Brainwave Verification System Integrating Passwords with EEG Templates for Online Identification and Authentication},
  author={Zhao, Hongze and Wang, Yijun and Gao, Xiaorong},
  journal={Pattern Recognition},
  volume={169},
  pages={112009},
  year={2026},
  publisher={Elsevier},
  doi={10.1016/j.patcog.2025.112009}
}

@article{arias2023consumer,
  title={Performance and Usability Evaluation of Brainwave Authentication Techniques with Consumer Devices},
  author={Arias-Cabarcos, Patricia and Fallahi, Matin and Habrich, Thilo and Schulze, Karen and Becker, Christian and Strufe, Thorsten},
  journal={ACM Transactions on Privacy and Security},
  volume={26},
  number={3},
  pages={26:1--26:36},
  year={2023},
  publisher={ACM New York, NY, USA},
  doi={10.1145/3579356}
}

@article{jurcak2007ten,
  title={10/20, 10/10, and 10/5 systems revisited: Their validity as relative head-surface-based positioning systems},
  author={Jurcak, Vit and Tsuzuki, Daisuke and Dan, Ippeita},
  journal={NeuroImage},
  volume={34},
  number={4},
  pages={1600--1611},
  year={2007},
  publisher={Elsevier},
  doi={10.1016/j.neuroimage.2006.09.024}
}

@article{chaurasia2024neuroidbench,
  title={NeuroIDBench: An open-source benchmark framework for the standardization of methodology in brainwave-based authentication research},
  author={Chaurasia, Avinash Kumar and Fallahi, Matin and Strufe, Thorsten and Terh{\"o}rst, Philipp and Cabarcos, Patricia Arias},
  journal={Journal of Information Security and Applications},
  volume={85},
  pages={103832},
  year={2024},
  publisher={Elsevier}
}

@article{chan2018challenges,
  title={Challenges and future perspectives on electroencephalogram-based biometrics in person recognition},
  author={Chan, Hui-Ling and Kuo, Po-Chih and Cheng, Chia-Yi and Chen, Yong-Sheng},
  journal={Frontiers in neuroinformatics},
  volume={12},
  pages={66},
  year={2018},
  publisher={Frontiers Media SA}
}

@article{fallahi2026advancing,
  title={Advancing brainwave-based biometrics: a large-scale, multi-session evaluation},
  author={Fallahi, Matin and Arias-Cabarcos, Patricia and Strufe, Thorsten},
  journal={IEEE Transactions on Biometrics, Behavior, and Identity Science},
  volume={8},
  number={3},
  pages={445--460},
  year={2026},
  publisher={IEEE},
  doi={10.1109/TBIOM.2026.3666044}
}

@article{hernandez2022eeg,
  title={EEG authentication system based on one-and multi-class machine learning classifiers},
  author={Hern{\'a}ndez-{\'A}lvarez, Luis and Barbierato, Elena and Caputo, Stefano and Mucchi, Lorenzo and Hern{\'a}ndez Encinas, Luis},
  journal={Sensors},
  volume={23},
  number={1},
  pages={186},
  year={2022},
  publisher={MDPI}
}

@article{al2025eeg,
  title={EEG-Based Authentication Across Various Event-Related Potentials (ERPs)},
  author={Al-Nafjan, Abeer and Alahaideb, Lamia and Aldayel, Mashael and Aljumah, Hessah},
  journal={Sensors},
  volume={25},
  number={16},
  pages={4962},
  year={2025},
  publisher={MDPI}
}

@inproceedings{fallahi2023brainnet,
  title={Brainnet: Improving brainwave-based biometric recognition with siamese networks},
  author={Fallahi, Matin and Strufe, Thorsten and Arias-Cabarcos, Patricia},
  booktitle={2023 IEEE International Conference on Pervasive Computing and Communications (PerCom)},
  pages={53--60},
  year={2023},
  organization={IEEE},
  doi={10.1109/PERCOM56429.2023.10099367}
}

@article{debie2021session,
  title={Session invariant EEG signatures using elicitation protocol fusion and convolutional neural network},
  author={Debie, Essam and Moustafa, Nour and Vasilakos, Athanasios},
  journal={IEEE Transactions on Dependable and Secure Computing},
  volume={19},
  number={4},
  pages={2488--2500},
  year={2021},
  publisher={IEEE}
}

@techreport{nist80063b2025,
  title={Digital Identity Guidelines: Authentication and Authenticator Management},
  author={{National Institute of Standards and Technology}},
  institution={National Institute of Standards and Technology},
  year={2025},
  url={https://pages.nist.gov/800-63-4/sp800-63b.html}
}

@misc{iso30107_1_2023,
  title={ISO/IEC 30107-1:2023 Information Technology --- Biometric Presentation Attack Detection --- Part 1: Framework},
  author={{International Organization for Standardization}},
  year={2023},
  url={https://www.iso.org/standard/83828.html},
  note={International Standard}
}

@inproceedings{nie2023patchtst,
  author       = {Yuqi Nie and
                  Nam H. Nguyen and
                  Phanwadee Sinthong and
                  Jayant Kalagnanam},
  title        = {A Time Series is Worth 64 Words: Long-term Forecasting with Transformers},
  booktitle    = {The Eleventh International Conference on Learning Representations,
                  {ICLR} 2023, Kigali, Rwanda, May 1-5, 2023},
  publisher    = {OpenReview.net},
  year         = {2023},
  url          = {https://openreview.net/forum?id=Jbdc0vTOcol},
  bibsource    = {dblp computer science bibliography, https://dblp.org}
}

@article{wang2024medformer,
  title={Medformer: A multi-granularity patching transformer for medical time-series classification},
  author={Wang, Yihe and Huang, Nan and Li, Taida and Yan, Yujun and Zhang, Xiang},
  journal={Advances in Neural Information Processing Systems},
  volume={37},
  pages={36314--36341},
  year={2024}
}

@inproceedings{press2022train,
  author       = {Ofir Press and
                  Noah A. Smith and
                  Mike Lewis},
  title        = {Train Short, Test Long: Attention with Linear Biases Enables Input
                  Length Extrapolation},
  booktitle    = {The Tenth International Conference on Learning Representations, {ICLR}
                  2022, Virtual Event, April 25-29, 2022},
  publisher    = {OpenReview.net},
  year         = {2022},
  url          = {https://openreview.net/forum?id=R8sQPpGCv0},
  bibsource    = {dblp computer science bibliography, https://dblp.org}
}

@inproceedings{khosla2020supervised,
  author       = {Prannay Khosla and
                  Piotr Teterwak and
                  Chen Wang and
                  Aaron Sarna and
                  Yonglong Tian and
                  Phillip Isola and
                  Aaron Maschinot and
                  Ce Liu and
                  Dilip Krishnan},
  editor       = {Hugo Larochelle and
                  Marc'Aurelio Ranzato and
                  Raia Hadsell and
                  Maria{-}Florina Balcan and
                  Hsuan{-}Tien Lin},
  title        = {Supervised Contrastive Learning},
  booktitle    = {Advances in Neural Information Processing Systems 33: Annual Conference
                  on Neural Information Processing Systems 2020, NeurIPS 2020, December
                  6-12, 2020, virtual},
  year         = {2020},
  url          = {https://proceedings.neurips.cc/paper/2020/hash/d89a66c7c80a29b1bdbab0f2a1a94af8-Abstract.html},
  bibsource    = {dblp computer science bibliography, https://dblp.org}
}

@inproceedings{wang2019multisimilarity,
  title={Multi-Similarity Loss with General Pair Weighting for Deep Metric Learning},
  author={Wang, Xun and Han, Xinge and Huang, Weilin and Dong, Dengke and Scott, Matthew R.},
  booktitle={Proceedings of the IEEE/CVF Conference on Computer Vision and Pattern Recognition},
  pages={5022--5030},
  year={2019},
  doi={10.1109/CVPR.2019.00516}
}

@inproceedings{harati2014tuh,
  title={The TUH EEG CORPUS: A big data resource for automated EEG interpretation},
  author={Harati, Amir and Lopez, Silvia and Obeid, I and Picone, J and Jacobson, MP and Tobochnik, S},
  booktitle={2014 IEEE signal processing in medicine and biology symposium (SPMB)},
  pages={1--5},
  year={2014},
  organization={IEEE}
}

@misc{ds005385:1.0.3,
  author = {Edmund Wascher AND Daniel Schneider AND Patrick D. Gajewski AND Stephan Getzmann},
  title = {Resting-state EEG data before and after cognitive activity across the adult lifespan and a 5-year follow-up},
  year = {2025},
  doi = {10.18112/openneuro.ds005385.v1.0.3},
  url = {https://openneuro.org/datasets/ds005385/versions/1.0.3},
  howpublished = {OpenNeuro dataset}
}

@article{getzmann2024resting,
  title={Resting-state EEG data before and after cognitive activity across the adult lifespan and a 5-year follow-up},
  author={Getzmann, Stephan and Gajewski, Patrick D and Schneider, Daniel and Wascher, Edmund},
  journal={Scientific Data},
  volume={11},
  number={1},
  pages={988},
  year={2024},
  publisher={Nature Publishing Group UK London}
}

@misc{ds005486:1.0.1,
  author = {Nahian S. Chowdhury AND Chuan Bi AND Andrew J. Furman AND Alan KI Chiang AND Patrick Skippen AND Emily Si AND Samantha K Millard AND Sarah M. Margerison AND Darrah Spies AND Michael L. Keaser AND Joyce T. Da Silva AND Shuo Chen AND Siobhan M. Schabrun AND David A. Seminowicz},
  title = {PREDICT},
  year = {2025},
  doi = {10.18112/openneuro.ds005486.v1.0.1},
  url = {https://openneuro.org/datasets/ds005486/versions/1.0.1},
  howpublished = {OpenNeuro dataset}
}

@article{seminowicz2020novel,
  title={A novel cortical biomarker signature for predicting pain sensitivity: protocol for the PREDICT longitudinal analytical validation study},
  author={Seminowicz, David A and Bilska, Katarzyna and Chowdhury, Nahian S and Skippen, Patrick and Millard, Samantha K and Chiang, Alan KI and Chen, Shuo and Furman, Andrew J and Schabrun, Siobhan M},
  journal={Pain Reports},
  volume={5},
  number={4},
  pages={e833},
  year={2020},
  publisher={LWW}
}

@article{chowdhury2025predicting,
  title={Predicting individual pain sensitivity using a novel cortical biomarker signature},
  author={Chowdhury, Nahian S and Bi, Chuan and Furman, Andrew J and Chiang, Alan KI and Skippen, Patrick and Si, Emily and Millard, Samantha K and Margerison, Sarah M and Spies, Darrah and Keaser, Michael L and others},
  journal={JAMA neurology},
  volume={82},
  number={3},
  pages={237--246},
  year={2025}
}

@misc{ds004395:2.0.0,
  author = {Michael J. Kahana AND Joseph H. Rudoler AND Lynn J. Lohnas AND Karl Healey AND Ada Aka AND Adam Broitman AND Elizabeth Crutchley AND Patrick Crutchley AND Kylie H. Alm AND Brandon S. Katerman AND Nicole E. Miller AND Joel R. Kuhn AND Yuxuan Li AND Nicole M. Long AND Jonathan Miller AND Madison D. Paron AND Jesse K. Pazdera AND Isaac Pedisich AND Christoph T. Weidemann},
  title = {Penn Electrophysiology of Encoding and Retrieval Study (PEERS)},
  year = {2023},
  doi = {10.18112/openneuro.ds004395.v2.0.0},
  url = {https://openneuro.org/datasets/ds004395/versions/2.0.0},
  howpublished = {OpenNeuro dataset}
}

@article{kahana2024penn,
  title={The Penn Electrophysiology of Encoding and Retrieval Study.},
  author={Kahana, Michael J and Lohnas, Lynn J and Healey, M Karl and Aka, Ada and Broitman, Adam W and Crutchley, Patrick and Crutchley, Elizabeth and Alm, Kylie H and Katerman, Brandon S and Miller, Nicole E and others},
  journal={Journal of Experimental Psychology: Learning, Memory, and Cognition},
  year={2024},
  publisher={American Psychological Association}
}

@article{albasri2019eeg,
  title={EEG dataset of Fusion relaxation and concentration moods},
  author={Albasri, A},
  journal={Mendeley Data, v1},
  year={2019},
  doi={10.17632/8c26dn6c7w.1},
  url={https://data.mendeley.com/datasets/8c26dn6c7w/1}
}

@article{long2014subsequent,
  title={Subsequent memory effect in intracranial and scalp EEG},
  author={Long, Nicole M. and Burke, John F. and Kahana, Michael J.},
  journal={NeuroImage},
  volume={84},
  pages={488--494},
  year={2014},
  doi={10.1016/j.neuroimage.2013.08.052}
}

@article{tenke2015posterior,
  title={Posterior EEG alpha at rest and during task performance: Comparison of current source density and field potential measures},
  author={Tenke, Craig E. and Kayser, J{\"u}rgen and Abraham, Karen and Alvarenga, Jorge E. and Bruder, Gerard E.},
  journal={International Journal of Psychophysiology},
  volume={97},
  number={3},
  pages={299--309},
  year={2015},
  doi={10.1016/j.ijpsycho.2015.05.011}
}

@article{bigdely2015prep,
  title={The PREP pipeline: standardized preprocessing for large-scale EEG analysis},
  author={Bigdely-Shamlo, Nima and Mullen, Tim and Kothe, Christian and Su, Kyung-Min and Robbins, Kay A.},
  journal={Frontiers in Neuroinformatics},
  volume={9},
  pages={16},
  year={2015},
  publisher={Frontiers Media SA},
  doi={10.3389/fninf.2015.00016}
}

@article{kessler2025preprocessing,
  title={How EEG preprocessing shapes decoding performance},
  author={Kessler, Roman and Enge, Alexander and Skeide, Michael A.},
  journal={Communications Biology},
  volume={8},
  pages={1039},
  year={2025},
  publisher={Springer Nature},
  doi={10.1038/s42003-025-08464-3}
}

@article{stergiadis2022personalized,
  title={A Personalized User Authentication System Based on EEG Signals},
  author={Stergiadis, Christos and Kostaridou, Vasiliki-Despoina and Veloudis, Simos and Kazis, Dimitrios and Klados, Manousos A.},
  journal={Sensors},
  volume={22},
  number={18},
  pages={6929},
  year={2022},
  publisher={MDPI},
  doi={10.3390/s22186929}
}

@article{lemke2024hans,
  title={Hans Berger and 100 years of the electroencephalogram: Insights into his life and his research on the ``electrencephalogram''},
  author={Lemke, Johannes R. and Kluger, Gerhard and Kr{\"a}mer, G{\"u}nter},
  journal={Clinical Epileptology},
  volume={37},
  number={Suppl 3},
  pages={112--119},
  year={2024},
  publisher={Springer},
  doi={10.1007/s10309-024-00704-6}
}

@article{sawangjai2020consumer,
  title={Consumer Grade EEG Measuring Sensors as Research Tools: A Review},
  author={Sawangjai, Phattarapong and Hompoonsup, Supanida and Leelaarporn, Pitshaporn and Kongwudhikunakorn, Supavit and Wilaiprasitporn, Theerawit},
  journal={IEEE Sensors Journal},
  volume={20},
  number={8},
  pages={3996--4024},
  year={2020},
  publisher={IEEE},
  doi={10.1109/JSEN.2019.2962874}
}

@inproceedings{vaswani2017attention,
  title={Attention Is All You Need},
  author={Vaswani, Ashish and Shazeer, Noam and Parmar, Niki and Uszkoreit, Jakob and Jones, Llion and Gomez, Aidan N. and Kaiser, {\L}ukasz and Polosukhin, Illia},
  booktitle={Advances in Neural Information Processing Systems},
  volume={30},
  year={2017}
}

@inproceedings{zerveas2021transformer,
  title={A Transformer-Based Framework for Multivariate Time Series Representation Learning},
  author={Zerveas, George and Jayaraman, Srideepika and Patel, Dhaval and Bhamidipaty, Anuradha and Eickhoff, Carsten},
  booktitle={Proceedings of the 27th ACM SIGKDD Conference on Knowledge Discovery \& Data Mining},
  pages={2114--2124},
  year={2021},
  doi={10.1145/3447548.3467401}
}

@inproceedings{zhou2021informer,
  title={Informer: Beyond Efficient Transformer for Long Sequence Time-Series Forecasting},
  author={Zhou, Haoyi and Zhang, Shanghang and Peng, Jieqi and Zhang, Shuai and Li, Jianxin and Xiong, Hui and Zhang, Wancai},
  booktitle={Proceedings of the AAAI Conference on Artificial Intelligence},
  volume={35},
  number={12},
  pages={11106--11115},
  year={2021}
}

@article{du2022etst,
  title={EEG Temporal--Spatial Transformer for Person Identification},
  author={Du, Yang and Xu, Yongling and Wang, Xiaoan and Liu, Li and Ma, Pengcheng},
  journal={Scientific Reports},
  volume={12},
  pages={14378},
  year={2022},
  doi={10.1038/s41598-022-18502-3}
}

@article{cai2023aitst,
  title={AITST---Affective EEG-Based Person Identification via Interrelated Temporal--Spatial Transformer},
  author={Cai, Honghua and Jin, Jiarui and Wang, Haoyu and Li, Liujiang and Huang, Yucui and Pan, Jiahui},
  journal={Pattern Recognition Letters},
  volume={174},
  pages={32--38},
  year={2023},
  doi={10.1016/j.patrec.2023.08.010}
}

@article{shao2026clt,
  title={An Interpretable Contrastive Learning Transformer for EEG-Based Person Identification},
  author={Shao, Xinghan and Chang, C. and Gan, John Q. and Wang, Haixian},
  journal={IEEE Transactions on Information Forensics and Security},
  volume={20},
  pages={5069--5082},
  year={2025},
  doi={10.1109/TIFS.2025.3570183}
}

@article{banville2026neuralbench,
  author       = {Hubert J. Banville and
                  St{\'{e}}phane d'Ascoli and
                  Simon Dahan and
                  J{\'{e}}r{\'{e}}my Rapin and
                  Marl{\`{e}}ne Careil and
                  Yohann Benchetrit and
                  Jarod L{\'{e}}vy and
                  Saarang Panchavati and
                  Antoine Ratouchniak and
                  Mingfang Zhang and
                  Elisa Cascardi and
                  Katelyn Begany and
                  Teon Brooks and
                  Jean{-}R{\'{e}}mi King},
  title        = {NeuralBench: {A} Unifying Framework to Benchmark NeuroAI Models},
  journal      = {CoRR},
  volume       = {abs/2605.08495},
  year         = {2026},
  url          = {https://doi.org/10.48550/arXiv.2605.08495},
  doi          = {10.48550/ARXIV.2605.08495},
  eprinttype   = {arXiv},
  eprint       = {2605.08495},
  bibsource    = {dblp computer science bibliography, https://dblp.org}
}

@article{elouahidi2025reve,
  author       = {Yassine El Ouahidi and
                  Jonathan Lys and
                  Philipp Th{\"{o}}lke and
                  Nicolas Farrugia and
                  Bastien Pasdeloup and
                  Vincent Gripon and
                  Karim Jerbi and
                  Giulia Lioi},
  title        = {{REVE:} {A} Foundation Model for {EEG} - Adapting to Any Setup with
                  Large-Scale Pretraining on 25,000 Subjects},
  journal      = {CoRR},
  volume       = {abs/2510.21585},
  year         = {2025},
  url          = {https://doi.org/10.48550/arXiv.2510.21585},
  doi          = {10.48550/ARXIV.2510.21585},
  eprinttype   = {arXiv},
  eprint       = {2510.21585},
  bibsource    = {dblp computer science bibliography, https://dblp.org}
}

@inproceedings{jiang2024labram,
  author       = {Wei{-}Bang Jiang and
                  Li{-}Ming Zhao and
                  Bao{-}Liang Lu},
  title        = {Large Brain Model for Learning Generic Representations with Tremendous
                  {EEG} Data in {BCI}},
  booktitle    = {The Twelfth International Conference on Learning Representations,
                  {ICLR} 2024, Vienna, Austria, May 7-11, 2024},
  publisher    = {OpenReview.net},
  year         = {2024},
  url          = {https://openreview.net/forum?id=QzTpTRVtrP},
  bibsource    = {dblp computer science bibliography, https://dblp.org}
}

@article{li2025enhancing,
  title={Enhancing EEG-Based Authentication With Transformer in Internet of Things},
  author={Li, Chunxue and Meng, Weizhi and Li, Wenjuan},
  journal={IEEE Transactions on Information Forensics and Security},
  volume={20},
  pages={7197--7210},
  year={2025},
  doi={10.1109/TIFS.2025.3586486}
}

@article{tangermann2012review,
  title={Review of the {BCI} Competition {IV}},
  author={Tangermann, Michael and M{\"u}ller, Klaus-Robert and Aertsen, Ad and Birbaumer, Niels and Braun, Christoph and Brunner, Clemens and Leeb, Robert and Mehring, Carsten and Miller, K. J. and Mueller-Putz, Gernot and Nolte, Guido and Pfurtscheller, Gert and Preissl, Hubert and Schalk, Gerwin and Schl{\"o}gl, Alois and Vidaurre, Carmen and Waldert, Stephan and Blankertz, Benjamin},
  journal={Frontiers in Neuroscience},
  volume={6},
  pages={55},
  year={2012},
  doi={10.3389/fnins.2012.00055}
}

@article{schalk2004bci2000,
  title={{BCI2000}: A General-Purpose Brain-Computer Interface ({BCI}) System},
  author={Schalk, Gerwin and McFarland, Dennis J. and Hinterberger, Thilo and Birbaumer, Niels and Wolpaw, Jonathan R.},
  journal={IEEE Transactions on Biomedical Engineering},
  volume={51},
  number={6},
  pages={1034--1043},
  year={2004},
  doi={10.1109/TBME.2004.827072}
}
\appendices

\section{Preprocessing Details}
\label{app:preprocessing}
Following common EEG preprocessing practice, we apply a
unified preprocessing pipeline across all datasets to
harmonize recordings collected with different EEG devices.
We first apply common average rereferencing to reduce
common-mode noise~\cite{kessler2025preprocessing,bigdely2015prep}.
We then band-pass filter the recordings to 1--45~Hz to remove
slow drift and high-frequency noise while preserving the EEG
frequency content most relevant for authentication
analysis~\cite{kessler2025preprocessing,stergiadis2022personalized}.
Next, we resample all recordings to 500~Hz to unify the
sampling rate across heterogeneous datasets and to align
with the sample resolution used in a recent multi-session
EEG-authentication study that serves as our main comparison
baseline~\cite{fallahi2026advancing}. Finally, we apply
amplitude-based quality control to remove segments that are
effectively flat or dominated by large artifacts.

\section{Effect of Enrollment and Verification Segment Count}
\label{app:session-protocol-evidence}

As a complementary protocol analysis,
Table~\ref{tab:session-protocol-summary-heatmap} varies the
number of enrollment sessions, the verification-session
scope, and the matched number of enrollment and
verification samples per session. Several consistent
patterns emerge. First, increasing the number of
samples on both sides generally lowers both EER
and FRR, with the largest gains occurring when
moving from very small sample counts to moderate
ones. Second, using two enrollment sessions is
consistently more effective than using only one,
especially at low sample counts, which suggests
that temporal diversity in template formation is
valuable beyond simply increasing the number of
segments. Third, expanding verification from only
the next session to all later sessions makes
the task harder overall, but the same scaling
pattern largely remains. These results therefore
support the view that NeuroShield benefits from
richer enrollment and verification evidence, while
also showing that session diversity is an important
part of that benefit.

\begin{table*}[!t]
\centering
\scriptsize
\renewcommand{\arraystretch}{1.10}
\setlength{\tabcolsep}{4.2pt}
\caption{Heatmap view of the session-aware authentication results on PEERS and FRC-EEG. For each protocol row, the PEERS metric block is colored according to that row\'s PEERS EER and the FRC-EEG metric block is colored according to that row\'s FRC-EEG EER. Greener rows indicate lower EER, yellow indicates intermediate EER, and redder rows indicate higher EER.}
\label{tab:session-protocol-summary-heatmap}
\begin{tabular}{|c|c|c|c|c|c|c|c|c|c|c|c|}
\hline
\multicolumn{4}{|c|}{\textbf{Protocol}} & \multicolumn{4}{c|}{\textbf{PEERS}} & \multicolumn{4}{c|}{\textbf{FRC-EEG}} \\
\hline
\textbf{Enr. Sess.} & \textbf{Ver. Scope} & \textbf{Samp.} & \textbf{Dur.} & \textbf{EER} & \textbf{FRR@1\%} & \textbf{FRR@0.1\%} & \textbf{FRR@0.01\%} & \textbf{EER} & \textbf{FRR@1\%} & \textbf{FRR@0.1\%} & \textbf{FRR@0.01\%} \\
\hline
\multirow{10}{*}{\textbf{1}} & \multirow{10}{*}{\textbf{1}} & \multirow{2}{*}{1} & 1 s & \cellcolor[RGB]{248,217,204}$10.0 \pm \text{\scriptsize 6.9}$ & \cellcolor[RGB]{248,217,204}47.0 & \cellcolor[RGB]{248,217,204}74.1 & \cellcolor[RGB]{248,217,204}88.0 & \cellcolor[RGB]{244,204,204}$23.7 \pm \text{\scriptsize 5.0}$ & \cellcolor[RGB]{244,204,204}81.4 & \cellcolor[RGB]{244,204,204}93.4 & \cellcolor[RGB]{244,204,204}97.2 \\
\cline{4-12}
  &   &   & 4 s & \cellcolor[RGB]{248,217,204}$6.7 \pm \text{\scriptsize 6.8}$ & \cellcolor[RGB]{248,217,204}32.0 & \cellcolor[RGB]{248,217,204}55.8 & \cellcolor[RGB]{248,217,204}71.8 & \cellcolor[RGB]{251,228,204}$16.3 \pm \text{\scriptsize 7.1}$ & \cellcolor[RGB]{251,228,204}68.1 & \cellcolor[RGB]{251,228,204}87.5 & \cellcolor[RGB]{251,228,204}94.3 \\
\cline{3-12}
  &   & \multirow{2}{*}{2} & 1 s & \cellcolor[RGB]{249,242,204}$5.6 \pm \text{\scriptsize 5.9}$ & \cellcolor[RGB]{249,242,204}29.5 & \cellcolor[RGB]{249,242,204}54.0 & \cellcolor[RGB]{249,242,204}71.1 & \cellcolor[RGB]{253,234,204}$14.4 \pm \text{\scriptsize 4.9}$ & \cellcolor[RGB]{253,234,204}65.6 & \cellcolor[RGB]{253,234,204}84.5 & \cellcolor[RGB]{253,234,204}92.0 \\
\cline{4-12}
  &   &   & 4 s & \cellcolor[RGB]{250,242,204}$4.0 \pm \text{\scriptsize 5.7}$ & \cellcolor[RGB]{250,242,204}20.5 & \cellcolor[RGB]{250,242,204}38.4 & \cellcolor[RGB]{250,242,204}51.7 & \cellcolor[RGB]{246,242,204}$10.3 \pm \text{\scriptsize 4.8}$ & \cellcolor[RGB]{246,242,204}50.9 & \cellcolor[RGB]{246,242,204}70.9 & \cellcolor[RGB]{246,242,204}78.4 \\
\cline{3-12}
  &   & \multirow{2}{*}{4} & 1 s & \cellcolor[RGB]{228,241,205}$3.3 \pm \text{\scriptsize 5.0}$ & \cellcolor[RGB]{228,241,205}19.0 & \cellcolor[RGB]{228,241,205}37.7 & \cellcolor[RGB]{228,241,205}51.0 & \cellcolor[RGB]{240,241,205}$9.1 \pm \text{\scriptsize 5.0}$ & \cellcolor[RGB]{240,241,205}44.1 & \cellcolor[RGB]{240,241,205}62.3 & \cellcolor[RGB]{240,241,205}73.1 \\
\cline{4-12}
  &   &   & 4 s & \cellcolor[RGB]{229,241,205}$2.5 \pm \text{\scriptsize 4.9}$ & \cellcolor[RGB]{229,241,205}14.9 & \cellcolor[RGB]{229,241,205}28.1 & \cellcolor[RGB]{229,241,205}36.7 & \cellcolor[RGB]{224,240,205}$5.8 \pm \text{\scriptsize 3.7}$ & \cellcolor[RGB]{224,240,205}27.8 & \cellcolor[RGB]{224,240,205}47.2 & \cellcolor[RGB]{224,240,205}57.0 \\
\cline{3-12}
  &   & \multirow{2}{*}{8} & 1 s & \cellcolor[RGB]{216,240,205}$2.1 \pm \text{\scriptsize 4.3}$ & \cellcolor[RGB]{216,240,205}14.8 & \cellcolor[RGB]{216,240,205}28.7 & \cellcolor[RGB]{216,240,205}37.1 & \cellcolor[RGB]{216,240,205}$4.2 \pm \text{\scriptsize 3.3}$ & \cellcolor[RGB]{216,240,205}21.1 & \cellcolor[RGB]{216,240,205}39.2 & \cellcolor[RGB]{216,240,205}49.9 \\
\cline{4-12}
  &   &   & 4 s & \cellcolor[RGB]{216,240,205}$1.7 \pm \text{\scriptsize 4.4}$ & \cellcolor[RGB]{216,240,205}12.7 & \cellcolor[RGB]{216,240,205}23.3 & \cellcolor[RGB]{216,240,205}28.1 & \cellcolor[RGB]{212,240,206}$3.3 \pm \text{\scriptsize 3.0}$ & \cellcolor[RGB]{212,240,206}12.7 & \cellcolor[RGB]{212,240,206}24.1 & \cellcolor[RGB]{212,240,206}32.5 \\
\cline{3-12}
  &   & \multirow{2}{*}{16} & 1 s & \cellcolor[RGB]{209,240,206}$1.5 \pm \text{\scriptsize 4.1}$ & \cellcolor[RGB]{209,240,206}12.0 & \cellcolor[RGB]{209,240,206}22.9 & \cellcolor[RGB]{209,240,206}28.3 & \cellcolor[RGB]{199,239,206}$0.8 \pm \text{\scriptsize 1.0}$ & \cellcolor[RGB]{199,239,206}1.5 & \cellcolor[RGB]{199,239,206}4.9 & \cellcolor[RGB]{199,239,206}10.2 \\
\cline{4-12}
  &   &   & 4 s & \cellcolor[RGB]{210,240,206}$1.3 \pm \text{\scriptsize 4.2}$ & \cellcolor[RGB]{210,240,206}11.5 & \cellcolor[RGB]{210,240,206}20.5 & \cellcolor[RGB]{210,240,206}23.6 & \cellcolor[RGB]{199,239,206}$0.7 \pm \text{\scriptsize 1.1}$ & \cellcolor[RGB]{199,239,206}1.5 & \cellcolor[RGB]{199,239,206}3.6 & \cellcolor[RGB]{199,239,206}5.6 \\
\hline
\multirow{10}{*}{\textbf{2}} & \multirow{10}{*}{\textbf{1}} & \multirow{2}{*}{1} & 1 s & \cellcolor[RGB]{254,240,204}$6.9 \pm \text{\scriptsize 4.8}$ & \cellcolor[RGB]{254,240,204}32.9 & \cellcolor[RGB]{254,240,204}61.1 & \cellcolor[RGB]{254,240,204}79.3 & \cellcolor[RGB]{249,220,204}$18.8 \pm \text{\scriptsize 4.2}$ & \cellcolor[RGB]{249,220,204}78.5 & \cellcolor[RGB]{249,220,204}92.3 & \cellcolor[RGB]{249,220,204}97.8 \\
\cline{4-12}
  &   &   & 4 s & \cellcolor[RGB]{254,240,204}$4.2 \pm \text{\scriptsize 4.3}$ & \cellcolor[RGB]{254,240,204}20.2 & \cellcolor[RGB]{254,240,204}42.6 & \cellcolor[RGB]{254,240,204}60.6 & \cellcolor[RGB]{253,242,204}$11.7 \pm \text{\scriptsize 4.9}$ & \cellcolor[RGB]{253,242,204}60.2 & \cellcolor[RGB]{253,242,204}79.3 & \cellcolor[RGB]{253,242,204}88.4 \\
\cline{3-12}
  &   & \multirow{2}{*}{2} & 1 s & \cellcolor[RGB]{225,240,205}$3.3 \pm \text{\scriptsize 3.6}$ & \cellcolor[RGB]{225,240,205}15.4 & \cellcolor[RGB]{225,240,205}38.4 & \cellcolor[RGB]{225,240,205}57.8 & \cellcolor[RGB]{249,242,204}$10.9 \pm \text{\scriptsize 3.9}$ & \cellcolor[RGB]{249,242,204}58.7 & \cellcolor[RGB]{249,242,204}78.2 & \cellcolor[RGB]{249,242,204}86.4 \\
\cline{4-12}
  &   &   & 4 s & \cellcolor[RGB]{225,240,205}$2.1 \pm \text{\scriptsize 3.0}$ & \cellcolor[RGB]{225,240,205}10.7 & \cellcolor[RGB]{225,240,205}25.3 & \cellcolor[RGB]{225,240,205}38.5 & \cellcolor[RGB]{229,241,205}$6.9 \pm \text{\scriptsize 4.5}$ & \cellcolor[RGB]{229,241,205}39.5 & \cellcolor[RGB]{229,241,205}62.2 & \cellcolor[RGB]{229,241,205}71.6 \\
\cline{3-12}
  &   & \multirow{2}{*}{4} & 1 s & \cellcolor[RGB]{209,240,206}$1.5 \pm \text{\scriptsize 2.3}$ & \cellcolor[RGB]{209,240,206}7.5 & \cellcolor[RGB]{209,240,206}21.5 & \cellcolor[RGB]{209,240,206}33.5 & \cellcolor[RGB]{231,241,205}$7.2 \pm \text{\scriptsize 4.6}$ & \cellcolor[RGB]{231,241,205}43.8 & \cellcolor[RGB]{231,241,205}59.2 & \cellcolor[RGB]{231,241,205}68.9 \\
\cline{4-12}
  &   &   & 4 s & \cellcolor[RGB]{210,240,206}$1.1 \pm \text{\scriptsize 2.3}$ & \cellcolor[RGB]{210,240,206}6.3 & \cellcolor[RGB]{210,240,206}14.5 & \cellcolor[RGB]{210,240,206}22.0 & \cellcolor[RGB]{216,240,205}$4.2 \pm \text{\scriptsize 3.9}$ & \cellcolor[RGB]{216,240,205}21.2 & \cellcolor[RGB]{216,240,205}37.0 & \cellcolor[RGB]{216,240,205}45.8 \\
\cline{3-12}
  &   & \multirow{2}{*}{8} & 1 s & \cellcolor[RGB]{203,239,206}$0.8 \pm \text{\scriptsize 1.8}$ & \cellcolor[RGB]{203,239,206}5.0 & \cellcolor[RGB]{203,239,206}14.1 & \cellcolor[RGB]{203,239,206}20.5 & \cellcolor[RGB]{213,240,205}$3.5 \pm \text{\scriptsize 3.0}$ & \cellcolor[RGB]{213,240,205}18.9 & \cellcolor[RGB]{213,240,205}37.6 & \cellcolor[RGB]{213,240,205}49.1 \\
\cline{4-12}
  &   &   & 4 s & \cellcolor[RGB]{203,239,206}$0.6 \pm \text{\scriptsize 1.6}$ & \cellcolor[RGB]{203,239,206}5.6 & \cellcolor[RGB]{203,239,206}10.6 & \cellcolor[RGB]{203,239,206}15.4 & \cellcolor[RGB]{204,239,206}$1.8 \pm \text{\scriptsize 3.0}$ & \cellcolor[RGB]{204,239,206}7.1 & \cellcolor[RGB]{204,239,206}13.3 & \cellcolor[RGB]{204,239,206}19.2 \\
\cline{3-12}
  &   & \multirow{2}{*}{16} & 1 s & \cellcolor[RGB]{199,239,206}$0.4 \pm \text{\scriptsize 1.2}$ & \cellcolor[RGB]{199,239,206}3.1 & \cellcolor[RGB]{199,239,206}7.7 & \cellcolor[RGB]{199,239,206}10.9 & \cellcolor[RGB]{201,239,206}$1.1 \pm \text{\scriptsize 1.9}$ & \cellcolor[RGB]{201,239,206}4.6 & \cellcolor[RGB]{201,239,206}10.2 & \cellcolor[RGB]{201,239,206}14.5 \\
\cline{4-12}
  &   &   & 4 s & \cellcolor[RGB]{198,239,206}$0.3 \pm \text{\scriptsize 1.0}$ & \cellcolor[RGB]{198,239,206}2.7 & \cellcolor[RGB]{198,239,206}5.3 & \cellcolor[RGB]{198,239,206}7.5 & \cellcolor[RGB]{198,239,206}$0.5 \pm \text{\scriptsize 1.4}$ & \cellcolor[RGB]{198,239,206}1.9 & \cellcolor[RGB]{198,239,206}4.3 & \cellcolor[RGB]{198,239,206}5.2 \\
\hline
\multirow{10}{*}{\textbf{1}} & \multirow{10}{*}{\textbf{all}} & \multirow{2}{*}{1} & 1 s & \cellcolor[RGB]{244,205,204}$12.2 \pm \text{\scriptsize 7.0}$ & \cellcolor[RGB]{244,205,204}54.3 & \cellcolor[RGB]{244,205,204}79.4 & \cellcolor[RGB]{244,205,204}91.3 & \cellcolor[RGB]{244,204,204}$23.7 \pm \text{\scriptsize 4.9}$ & \cellcolor[RGB]{244,204,204}84.7 & \cellcolor[RGB]{244,204,204}95.7 & \cellcolor[RGB]{244,204,204}98.9 \\
\cline{4-12}
  &   &   & 4 s & \cellcolor[RGB]{244,204,204}$8.5 \pm \text{\scriptsize 6.4}$ & \cellcolor[RGB]{244,204,204}39.1 & \cellcolor[RGB]{244,204,204}65.1 & \cellcolor[RGB]{244,204,204}81.0 & \cellcolor[RGB]{251,230,204}$15.8 \pm \text{\scriptsize 6.1}$ & \cellcolor[RGB]{251,230,204}70.7 & \cellcolor[RGB]{251,230,204}89.3 & \cellcolor[RGB]{251,230,204}96.6 \\
\cline{3-12}
  &   & \multirow{2}{*}{2} & 1 s & \cellcolor[RGB]{252,232,204}$8.0 \pm \text{\scriptsize 6.1}$ & \cellcolor[RGB]{252,232,204}37.6 & \cellcolor[RGB]{252,232,204}64.4 & \cellcolor[RGB]{252,232,204}80.9 & \cellcolor[RGB]{252,233,204}$14.9 \pm \text{\scriptsize 5.7}$ & \cellcolor[RGB]{252,233,204}70.2 & \cellcolor[RGB]{252,233,204}87.5 & \cellcolor[RGB]{252,233,204}93.8 \\
\cline{4-12}
  &   &   & 4 s & \cellcolor[RGB]{252,232,204}$5.9 \pm \text{\scriptsize 5.7}$ & \cellcolor[RGB]{252,232,204}26.7 & \cellcolor[RGB]{252,232,204}50.5 & \cellcolor[RGB]{252,232,204}67.3 & \cellcolor[RGB]{247,242,204}$10.4 \pm \text{\scriptsize 6.0}$ & \cellcolor[RGB]{247,242,204}50.8 & \cellcolor[RGB]{247,242,204}71.9 & \cellcolor[RGB]{247,242,204}82.1 \\
\cline{3-12}
  &   & \multirow{2}{*}{4} & 1 s & \cellcolor[RGB]{248,242,204}$5.6 \pm \text{\scriptsize 5.6}$ & \cellcolor[RGB]{248,242,204}26.1 & \cellcolor[RGB]{248,242,204}50.1 & \cellcolor[RGB]{248,242,204}67.2 & \cellcolor[RGB]{243,241,204}$9.6 \pm \text{\scriptsize 4.7}$ & \cellcolor[RGB]{243,241,204}51.2 & \cellcolor[RGB]{243,241,204}71.5 & \cellcolor[RGB]{243,241,204}81.2 \\
\cline{4-12}
  &   &   & 4 s & \cellcolor[RGB]{249,242,204}$4.4 \pm \text{\scriptsize 5.3}$ & \cellcolor[RGB]{249,242,204}19.8 & \cellcolor[RGB]{249,242,204}39.9 & \cellcolor[RGB]{249,242,204}55.0 & \cellcolor[RGB]{230,241,205}$7.0 \pm \text{\scriptsize 4.5}$ & \cellcolor[RGB]{230,241,205}33.1 & \cellcolor[RGB]{230,241,205}53.0 & \cellcolor[RGB]{230,241,205}62.5 \\
\cline{3-12}
  &   & \multirow{2}{*}{8} & 1 s & \cellcolor[RGB]{237,241,205}$4.4 \pm \text{\scriptsize 5.4}$ & \cellcolor[RGB]{237,241,205}20.2 & \cellcolor[RGB]{237,241,205}41.5 & \cellcolor[RGB]{237,241,205}56.1 & \cellcolor[RGB]{223,240,205}$5.6 \pm \text{\scriptsize 3.4}$ & \cellcolor[RGB]{223,240,205}31.2 & \cellcolor[RGB]{223,240,205}49.6 & \cellcolor[RGB]{223,240,205}59.3 \\
\cline{4-12}
  &   &   & 4 s & \cellcolor[RGB]{238,241,205}$3.6 \pm \text{\scriptsize 5.1}$ & \cellcolor[RGB]{238,241,205}16.2 & \cellcolor[RGB]{238,241,205}33.8 & \cellcolor[RGB]{238,241,205}46.7 & \cellcolor[RGB]{215,240,205}$4.0 \pm \text{\scriptsize 3.1}$ & \cellcolor[RGB]{215,240,205}16.4 & \cellcolor[RGB]{215,240,205}31.6 & \cellcolor[RGB]{215,240,205}38.6 \\
\cline{3-12}
  &   & \multirow{2}{*}{16} & 1 s & \cellcolor[RGB]{231,241,205}$3.7 \pm \text{\scriptsize 5.2}$ & \cellcolor[RGB]{231,241,205}16.9 & \cellcolor[RGB]{231,241,205}36.1 & \cellcolor[RGB]{231,241,205}47.6 & \cellcolor[RGB]{211,240,206}$3.2 \pm \text{\scriptsize 3.2}$ & \cellcolor[RGB]{211,240,206}12.7 & \cellcolor[RGB]{211,240,206}26.2 & \cellcolor[RGB]{211,240,206}35.1 \\
\cline{4-12}
  &   &   & 4 s & \cellcolor[RGB]{232,241,205}$3.3 \pm \text{\scriptsize 5.0}$ & \cellcolor[RGB]{232,241,205}14.4 & \cellcolor[RGB]{232,241,205}29.9 & \cellcolor[RGB]{232,241,205}40.5 & \cellcolor[RGB]{205,239,206}$2.0 \pm \text{\scriptsize 3.4}$ & \cellcolor[RGB]{205,239,206}6.4 & \cellcolor[RGB]{205,239,206}9.2 & \cellcolor[RGB]{205,239,206}12.7 \\
\hline
\multirow{10}{*}{\textbf{2}} & \multirow{10}{*}{\textbf{all}} & \multirow{2}{*}{1} & 1 s & \cellcolor[RGB]{251,228,204}$8.9 \pm \text{\scriptsize 5.0}$ & \cellcolor[RGB]{251,228,204}42.6 & \cellcolor[RGB]{251,228,204}69.9 & \cellcolor[RGB]{251,228,204}85.8 & \cellcolor[RGB]{247,216,204}$20.1 \pm \text{\scriptsize 4.4}$ & \cellcolor[RGB]{247,216,204}81.3 & \cellcolor[RGB]{247,216,204}93.8 & \cellcolor[RGB]{247,216,204}98.0 \\
\cline{4-12}
  &   &   & 4 s & \cellcolor[RGB]{251,228,204}$6.0 \pm \text{\scriptsize 4.4}$ & \cellcolor[RGB]{251,228,204}28.5 & \cellcolor[RGB]{251,228,204}53.7 & \cellcolor[RGB]{251,228,204}71.6 & \cellcolor[RGB]{254,240,204}$12.8 \pm \text{\scriptsize 5.8}$ & \cellcolor[RGB]{254,240,204}63.6 & \cellcolor[RGB]{254,240,204}82.6 & \cellcolor[RGB]{254,240,204}87.6 \\
\cline{3-12}
  &   & \multirow{2}{*}{2} & 1 s & \cellcolor[RGB]{245,241,204}$5.4 \pm \text{\scriptsize 3.9}$ & \cellcolor[RGB]{245,241,204}26.2 & \cellcolor[RGB]{245,241,204}51.8 & \cellcolor[RGB]{245,241,204}70.9 & \cellcolor[RGB]{255,241,204}$12.5 \pm \text{\scriptsize 5.2}$ & \cellcolor[RGB]{255,241,204}62.9 & \cellcolor[RGB]{255,241,204}82.3 & \cellcolor[RGB]{255,241,204}90.1 \\
\cline{4-12}
  &   &   & 4 s & \cellcolor[RGB]{246,242,204}$3.9 \pm \text{\scriptsize 3.5}$ & \cellcolor[RGB]{246,242,204}18.0 & \cellcolor[RGB]{246,242,204}38.0 & \cellcolor[RGB]{246,242,204}55.2 & \cellcolor[RGB]{235,241,205}$8.1 \pm \text{\scriptsize 5.5}$ & \cellcolor[RGB]{235,241,205}44.8 & \cellcolor[RGB]{235,241,205}64.2 & \cellcolor[RGB]{235,241,205}72.6 \\
\cline{3-12}
  &   & \multirow{2}{*}{4} & 1 s & \cellcolor[RGB]{228,241,205}$3.7 \pm \text{\scriptsize 3.3}$ & \cellcolor[RGB]{228,241,205}17.0 & \cellcolor[RGB]{228,241,205}37.7 & \cellcolor[RGB]{228,241,205}55.6 & \cellcolor[RGB]{237,241,205}$8.4 \pm \text{\scriptsize 4.7}$ & \cellcolor[RGB]{237,241,205}44.7 & \cellcolor[RGB]{237,241,205}62.8 & \cellcolor[RGB]{237,241,205}70.8 \\
\cline{4-12}
  &   &   & 4 s & \cellcolor[RGB]{229,241,205}$2.9 \pm \text{\scriptsize 3.0}$ & \cellcolor[RGB]{229,241,205}12.8 & \cellcolor[RGB]{229,241,205}28.2 & \cellcolor[RGB]{229,241,205}42.4 & \cellcolor[RGB]{223,240,205}$5.5 \pm \text{\scriptsize 5.0}$ & \cellcolor[RGB]{223,240,205}28.0 & \cellcolor[RGB]{223,240,205}41.9 & \cellcolor[RGB]{223,240,205}47.8 \\
\cline{3-12}
  &   & \multirow{2}{*}{8} & 1 s & \cellcolor[RGB]{221,240,205}$2.8 \pm \text{\scriptsize 3.0}$ & \cellcolor[RGB]{221,240,205}12.7 & \cellcolor[RGB]{221,240,205}29.8 & \cellcolor[RGB]{221,240,205}44.3 & \cellcolor[RGB]{223,240,205}$5.5 \pm \text{\scriptsize 4.1}$ & \cellcolor[RGB]{223,240,205}31.8 & \cellcolor[RGB]{223,240,205}46.7 & \cellcolor[RGB]{223,240,205}55.2 \\
\cline{4-12}
  &   &   & 4 s & \cellcolor[RGB]{222,240,205}$2.4 \pm \text{\scriptsize 2.9}$ & \cellcolor[RGB]{222,240,205}10.6 & \cellcolor[RGB]{222,240,205}23.1 & \cellcolor[RGB]{222,240,205}34.6 & \cellcolor[RGB]{212,240,206}$3.3 \pm \text{\scriptsize 4.4}$ & \cellcolor[RGB]{212,240,206}18.2 & \cellcolor[RGB]{212,240,206}26.3 & \cellcolor[RGB]{212,240,206}33.5 \\
\cline{3-12}
  &   & \multirow{2}{*}{16} & 1 s & \cellcolor[RGB]{217,240,205}$2.4 \pm \text{\scriptsize 2.9}$ & \cellcolor[RGB]{217,240,205}10.8 & \cellcolor[RGB]{217,240,205}24.6 & \cellcolor[RGB]{217,240,205}36.0 & \cellcolor[RGB]{209,240,206}$2.7 \pm \text{\scriptsize 3.5}$ & \cellcolor[RGB]{209,240,206}14.1 & \cellcolor[RGB]{209,240,206}25.2 & \cellcolor[RGB]{209,240,206}30.6 \\
\cline{4-12}
  &   &   & 4 s & \cellcolor[RGB]{217,240,205}$2.2 \pm \text{\scriptsize 2.8}$ & \cellcolor[RGB]{217,240,205}9.6 & \cellcolor[RGB]{217,240,205}19.9 & \cellcolor[RGB]{217,240,205}28.4 & \cellcolor[RGB]{205,239,206}$2.0 \pm \text{\scriptsize 3.4}$ & \cellcolor[RGB]{205,239,206}8.5 & \cellcolor[RGB]{205,239,206}12.6 & \cellcolor[RGB]{205,239,206}15.1 \\
\hline
\end{tabular}
\end{table*}

\section{Additional Spatial Ablation Analyses}
\label{app:spatial-ablations}
\label{app:regional-ablation}
\label{app:region-cross-region}

\begin{figure*}[!t]
    \centering
    \includegraphics[width=0.98\textwidth]{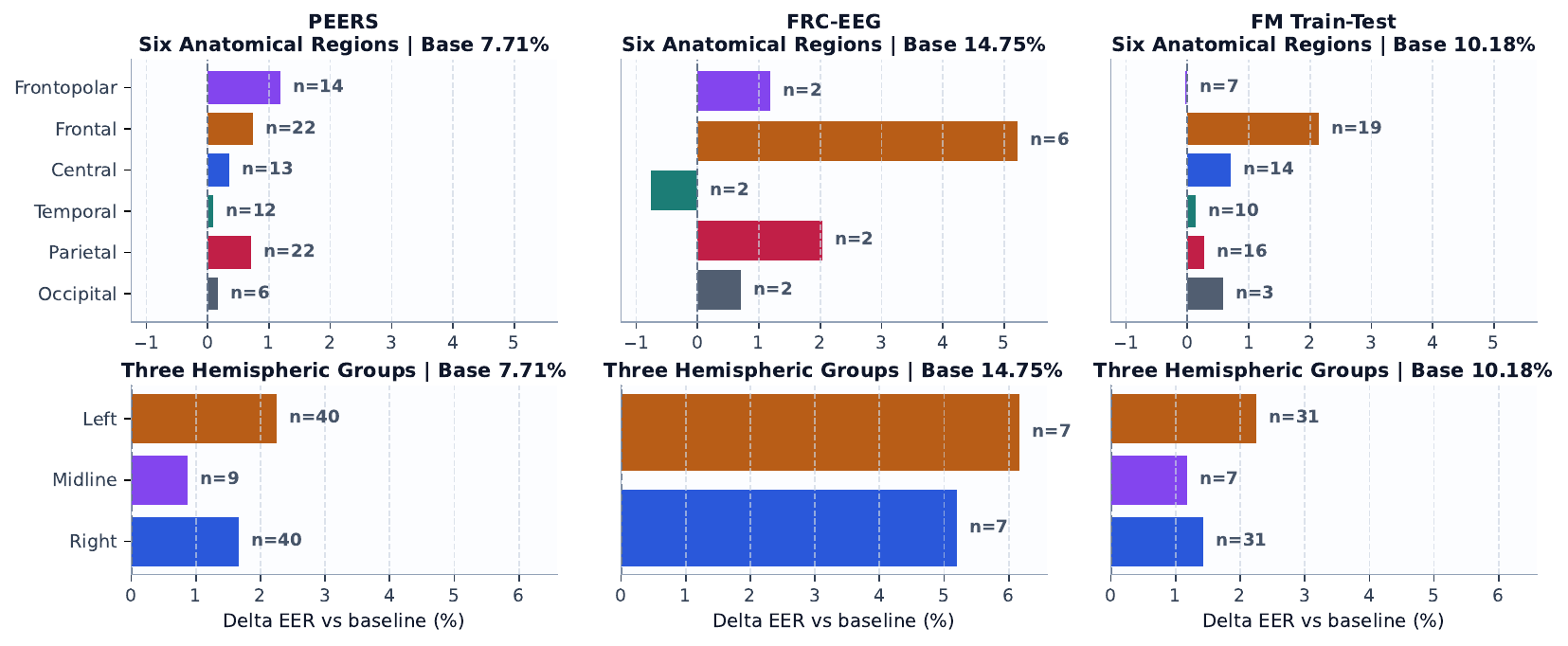}
    \caption{Regional ablation analysis across PEERS, FRC-EEG, and FM Train-Test. The top row uses a six-region anatomical grouping, and the bottom row uses a three-part hemisphere grouping. Each bar reports the change in EER relative to the corresponding full-channel baseline after removing that region, and the figure also indicates how many template channels were removed in each condition.}
    \label{fig:regional_analysis_overview}
\end{figure*}

This appendix collects the supplementary spatial analyses
referenced in the main text. Figure~\ref{fig:regional_analysis_overview}
provides the full region-level ablation view across PEERS,
FRC-EEG, and FM Train-Test. The top row uses a six-region
anatomical grouping, and the bottom row uses a coarser
hemisphere grouping.

\begin{figure*}[!t]
    \centering
    \includegraphics[width=0.96\textwidth]{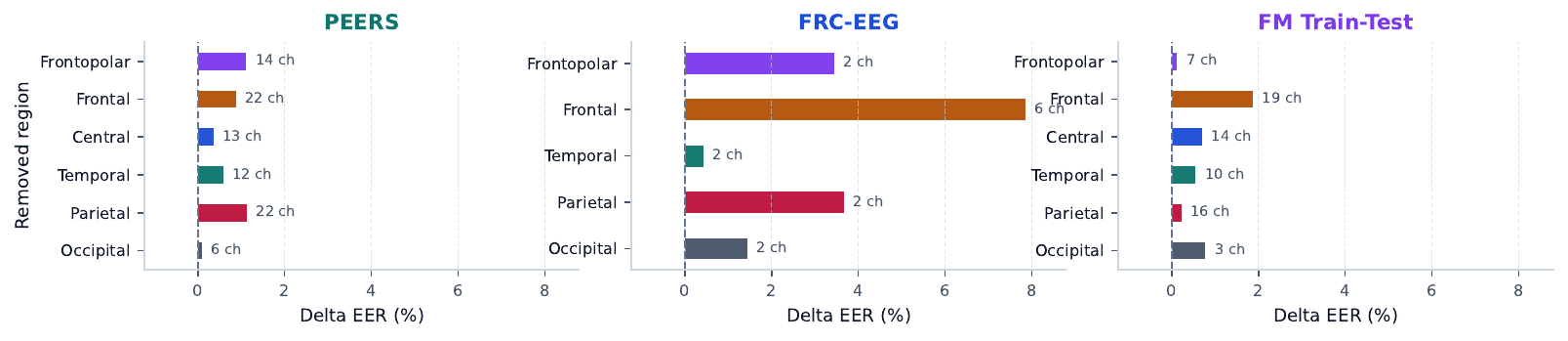}
    \caption{Authentication under removal of one anatomical region at verification across PEERS, FRC-EEG, and FM Train-Test. Each bar reports the change in EER relative to the full-channel baseline, and annotations show the number of removed template channels.}
    \label{fig:region_missing_multidataset}
\end{figure*}

As a stricter complement to the random missing-channel
analysis in Section~\ref{sec:missing-channels},
Figure~\ref{fig:region_missing_multidataset} removes one
anatomical region at verification and reports the resulting
change in EER relative to the full-channel baseline.
Removing one region is naturally more disruptive in sparser
montages, and the strongest effect appears in FRC-EEG,
where removing the frontal region causes the largest EER
increase.

\begin{figure*}[!t]
    \centering
    \includegraphics[width=0.96\textwidth]{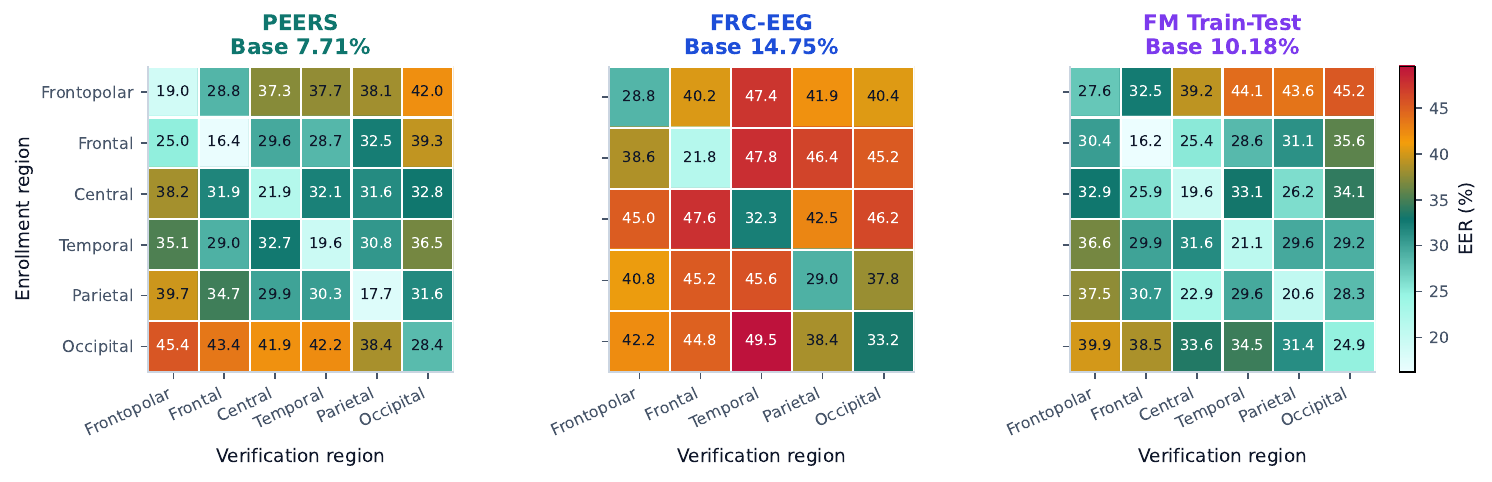}
    \caption{Cross-region authentication across PEERS, FRC-EEG, and FM Train-Test. Enrollment and verification are each restricted to a single anatomical region, and each cell reports the resulting EER.}
    \label{fig:cross_region_authentication_multidataset}
\end{figure*}

Figure~\ref{fig:cross_region_authentication_multidataset}
then restricts enrollment and verification to different
single anatomical regions. Matching enrollment and
verification within the same region is consistently
better than cross-region pairing. Although NeuroShield
is not explicitly trained for cross-region
authentication, the results indicate that some identity
information is shared across regions; however, this
transfer is weaker in FRC-EEG, likely because of its
smaller number of channels. Taken together, these
results show that, beyond robustness to modest random
missing channels, strong authentication still depends
on distributed spatial coverage rather than isolated or
mismatched scalp regions.

\end{document}